%% file: main.tex
\documentclass[numbers,10pt]{main}

\usepackage[utf8]{inputenc}

\input{preamble}

\makeatletter
\newcommand{\disablemaintoc}{%
  \let\orig@addcontentsline\addcontentsline
  \renewcommand{\addcontentsline}[3]{}%
}

\newcommand{\enablemaintoc}{%
  \let\addcontentsline\orig@addcontentsline
}
\makeatother

\usepackage[T1]{fontenc}
\usepackage{amsfonts}
\usepackage{amsthm}
\usepackage{amsmath}
\usepackage{nicefrac}
\usepackage{url}
\usepackage{array}
\usepackage{fontawesome5}
\usepackage{amssymb}

\renewcommand{\paragraph}[1]{\sansbf{#1}}

\definecolor{takeawaycolor}{RGB}{70, 70, 180}
\definecolor{boxbg}{RGB}{240, 240, 250}
\definecolor{rlratio}{HTML}{2C5282}

\newtcolorbox{takeaway}{%
  enhanced,
  rounded corners,
  breakable,
  colback=boxbg,
  colframe=boxbg,
  boxrule=0pt,
  left=10pt,
  right=10pt,
  top=5pt,
  bottom=5pt,
  toptitle=1pt,
  bottomtitle=1pt,
  arc=12pt,
  before={\vspace{0pt}},
  after={\vspace{0pt}}
}

\definecolor{lawgold}{HTML}{C58A16}
\definecolor{lawcream}{HTML}{FFF9EC}
\definecolor{lawtitlebg}{HTML}{FBECC8}
\definecolor{lawdark}{HTML}{6F4A08}

\newtcolorbox{scalinglaw}{
    enhanced,
    breakable,
    colback=lawcream,
    colframe=lawgold,
    colbacktitle=lawtitlebg,
    coltitle=lawdark,
    fonttitle=\bfseries,
    title={Joint Pretraining--RL Scaling Law},
    boxrule=0.7pt,
    titlerule=0.45pt,
    arc=1.8mm,
    left=3.5mm,
    right=3.5mm,
    top=2mm,
    bottom=2mm,
    toptitle=1.4mm,
    bottomtitle=1.4mm,
    before skip=8pt,
    after skip=8pt,
    drop fuzzy shadow=black!8
}

\title{Harnessing Domain Specialists in Multimodal Mixture-of-Experts for Efficient Adaptation 
\vspace{4mm}}

\author[\star]{Damiano Marsili}
\author[\star]{Raphi Kang}
\author[]{Aditya Mehta}
\author[]{\\Pietro Perona}
\author[]{Georgia Gkioxari}

\contribution[\ensuremath{\star}]{Equal contribution}

\affiliation{\textcolor[HTML]{FF6C0C}{\textbf{California Institute of Technology}}}

\theoremstyle{plain}

\theoremstyle{definition}

\theoremstyle{remark}

\begin{document}
\disablemaintoc

\input{sections/abstract}
\maketitle

\input{sections/intro}
\input{sections/related}
\input{sections/em_structure}
\input{sections/experiments}

\input{sections/conclusion}

\section*{Acknowledgments}

We thank Aadarsh Sahoo, Ilona Demler, and Ziqi Ma for their feedback on the project.
This project was funded by the Packard Fellowship, the Powell Foundation, an Amazon Research Award, a Google Faculty Award and a Meta Research Grant. We also thank Google for providing us with Gemini credits.
R.K. was supported by the NSF Graduate Research Fellowship Program under Grant No. 2139433.

\bibliographystyle{plainnat}
\small\bibliography{main}

\newpage
\appendix
\input{sections/appendix}

\end{document}

%% file: preamble.tex
\usepackage[dvipsnames]{xcolor}

\newcommand{\mypar}[1]{\vspace{1mm}\noindent\textbf{#1}}

\newcommand{\gain}[1]{%
  \makebox[2.5em][l]{%
    \textcolor{green!50!black}{\scriptsize~(#1)}%
  }%
}

\usepackage{booktabs}
\usepackage{multirow}
\usepackage{hyperref}
\usepackage{url}
\usepackage{wrapfig}
\usepackage{graphicx}
\usepackage[most]{tcolorbox}
\usepackage{xcolor}

\definecolor{findingblue}{HTML}{EEF6FF}
\definecolor{findingborder}{HTML}{3B82F6}

\newcommand{\qwen}{\textsc{Qwen3-VL-30B-A3B}\xspace}
\newcommand{\gemma}{\textsc{Gemma-4 26B-A4B}\xspace}
\newcommand{\kimi}{\textsc{Kimi-VL-A3B}\xspace}

\def\expertlens{\texttt{ExpertLens}\xspace}

\newtcbtheorem[number within=section]{finding}{Finding}{
  enhanced,
  breakable,
  colback=findingblue!6,
  colframe=findingborder!9,
  coltitle=black,
  fonttitle=\bfseries,
  boxrule=0.7pt,
  arc=2mm,
  left=2mm,
  right=2mm,
  top=1mm,
  bottom=1mm
}{find}

\usepackage{tabularx}
\usepackage{nicefrac}             % 1/2 etc.
\usepackage{graphicx, subcaption} % figures
\usepackage{enumitem}             % compact lists
\usepackage{array, tabularx}      % advanced tables
\newcolumntype{L}{>{\raggedright\arraybackslash}X} % define L column type
\usepackage{fvextra}              % nicer verbatim
\usepackage[most]{tcolorbox}      % call-outs
\usepackage{xspace}               % smart spaces
\usepackage{bxcoloremoji}
\usepackage{graphicx}
\usepackage{enumitem}
\usepackage{caption}
\usepackage{xspace}
\usepackage{fvextra}
\usepackage{cleveref}
\usepackage{float} 
\usepackage{subcaption}
\usepackage{booktabs}
\usepackage{adjustbox}
\usepackage{multirow}
\usepackage{siunitx} % for \num{} if desired
\usepackage{bbm}
\tcbuselibrary{listings,breakable,skins}
\usepackage{CJKutf8}

\def\tin{\textsc{TinyImageNet}\xspace}
\def\medmnist{\textsc{MedMnist}\xspace}
\def\flowers{\textsc{Flowers102}\xspace}
\def\breasthist{\textsc{BreastHistopathology}\xspace}
\def\cosyn{\textsc{CoSyn}\xspace}
\def\pathvqa{\textsc{PathVqa}\xspace}
\def\vqarad{\textsc{VqaRad}\xspace}
\def\slake{\textsc{Slake}\xspace}
\def\mathvista{\textsc{MathVista}\xspace}
\def\mathvision{\textsc{MathVision}\xspace}
\def\mmk{\textsc{Mmk12}\xspace}
\def\hrvqa{\textsc{HrVqa}\xspace}
\def\rsvlmqa{\textsc{RsVlmQa}\xspace}
\def\rsvqalr{\textsc{RsVqaLr}\xspace}
\def\rsvqahr{\textsc{RsVqaHr}\xspace}

\def\scienceqa{\textsc{ScienceQA}\xspace}
\def\aiid{\textsc{AI2D}\xspace}
\def\geomverse{\textsc{GeomVerse}\xspace}
\def\charxiv{\textsc{CharXiv}\xspace}

\def\medqa{\textsc{MedQA}\xspace}
\def\pubmedqa{\textsc{PubMedQA}\xspace}
\def\medmcqa{\textsc{MedMCQA}\xspace}
\def\gsm{\textsc{GSM8K}\xspace}
\def\metamath{\textsc{MetaMathQA}\xspace}
\def\arcchallenge{\textsc{ARC-Challenge}\xspace}
\def\openbookqa{\textsc{OpenBookQA}\xspace}
\def\mmlu{\textsc{MMLU}\xspace}
\def\triviaqa{\textsc{TriviaQA}\xspace}
\def\bbh{\textsc{BBH}\xspace}
\def\finqa{\textsc{FinQA}\xspace}
\def\legalbench{\textsc{LegalBench}\xspace}

\usepackage{tocloft}

\makeatletter
\DeclareRobustCommand\onedot{\futurelet\@let@token\@onedot}
\def\@onedot{\ifx\@let@token.\else.\null\fi\xspace}

\makeatother

\renewcommand{\paragraph}[1]{\textbf{#1}}

\usepackage{xcolor}
\usepackage{amsmath}
\usepackage{wrapfig}

\definecolor{promptblue}{RGB}{25, 84, 166}
\definecolor{thinkgreen}{RGB}{34, 139, 34}
\definecolor{thinktag}{RGB}{110, 65, 30}  % Darker, richer brown
\definecolor{thinkbg}{RGB}{252, 248, 243}  % Light cream
\definecolor{boxbg}{RGB}{252, 252, 252}

\tcbset{
    mainbox/.style={
        colback=boxbg,
        colframe=gray!40,
        arc=2mm,
        boxrule=0.8pt,
        left=12pt,
        right=12pt,
        top=12pt,
        bottom=12pt,
        fonttitle=\bfseries\large,
        coltitle=white,
        colbacktitle=gray!60
    }
}

%% file: sections/abstract.tex
% --- Abstract ---
\abstract{
\vspace{3mm}

Mixture-of-Experts (MoE) architectures scale model capacity through sparse computation, routing each token through only a small subset of experts. In this work, we explore whether this sparsity gives rise to emergent intrinsic organization in multimodal MoEs. We find that experts develop strong semantic specialization across modalities and domains despite not being explicitly trained for modularity. Building on this structure, we introduce \expertlens, a data-free method that identifies domain-specialized experts directly from pretrained model weights by decoding router weights into semantically meaningful vocabulary tokens. We leverage this specialization for efficient multimodal adaptation by selectively fine-tuning experts relevant to a target domain. Across math, medical, and remote sensing tasks, \expertlens matches or surpasses full fine-tuning while updating only $21.7$--$47.0\%$ of model parameters and achieving a $4.0\times$ average training speedup, and outperforms LoRA in both adaptation performance and training efficiency. These results show that sparsity introduced for efficiency can give rise to semantic modularity that is directly useful for efficient adaptation.

\vspace{3mm}

\faGlobe\ \textbf{Project Page}: \href{https://glab-caltech.github.io/expertlens}{glab-caltech.github.io/expertlens} \\
\faGithub\ \textbf{Code}: \href{https://github.com/damianomarsili/ExpertLens}{github.com/damianomarsili/ExpertLens}

\vspace{2mm}
}

%% file: sections/intro.tex
\section{Introduction}
Mixture-of-Experts (MoE) architectures have become a key ingredient in scaling modern language and multimodal models~\citep{gemma4, kimivl}. By replacing dense feed-forward layers with many experts while activating only a small subset for each token, MoEs decouple model capacity from per-token computation~\citep{shazeer2017, mcgill2017deciding,  fedus2022switch}. This sparsity has enabled models to scale to enormous sizes while remaining computationally tractable~\citep{team2026kimi, inkling}.

Sparsity in MoEs was introduced primarily for efficiency, not organization~\citep{lepikhin2020gshard}. Experts are learned jointly, with a router selecting a subset per token. Yet consistent routing patterns could induce implicit modularity, with experts specializing to different kinds of information. Such structure has been studied in text-only MoEs~\citep{antoine2025part, li_understanding_2026}, but only recently in multimodal models~\citep{xu2026seeing, zhang2026beyond}, where the same expert layers process both visual and textual representations. We therefore ask whether sparse multimodal MoEs are semantically organized: do experts specialize by modality, domain, and visual concept?

In this work, we find that multimodal MoEs develop striking semantic specialization without explicitly training for it. Across three large multimodal MoEs, experts exhibit strong modality specialization, preferentially processing visual or textual tokens (\S\ref{sec:expertspecialization_modality}). Beyond modality, visual representations and routing organize strongly by semantic domain: medical, math, and remote sensing imagery form distinct structures not explained by low-level properties such as color or spatial position (\S\ref{subsec:domain_specificity}). This extends to finer semantics, with visually dissimilar but semantically related object classes recruiting overlapping expert sets (\S\ref{subsec:semantic_specialization}). Thus, sparsity introduced for computational efficiency gives rise to an emergent semantic decomposition of multimodal MoEs.

This emergent structure raises a practical question: given an MoE, how can we identify which experts specialize in a given domain? To this end, we introduce \expertlens, a data-free method that recovers expert semantics directly from pretrained weights by decoding router weights into semantically meaningful vocabulary tokens (\S\ref{subsec:expertlens}). \expertlens enables identification of experts relevant to a domain without requiring domain-specific examples.

We next ask whether this emergent specialization can enable more efficient multimodal adaptation (\S\ref{sec:experiments}). Using \expertlens, we selectively adapt experts relevant to each domain while freezing the remaining expert MLPs. Across sequential adaptation to math, medical, and remote sensing tasks, this strategy matches or surpasses full fine-tuning while updating only $21.7$--$47.0\%$ of model parameters and achieving a $4.0\times$ average speedup. \expertlens also outperforms LoRA in both adaptation performance and training speed ($2.9\times$ faster). Thus, the emergent specialization of multimodal MoEs is not only interpretable, but directly useful for efficient adaptation.

In summary, our contributions are:
\begin{itemize}
    \item We uncover emergent semantic structure in multimodal MoEs, showing specialization across modalities, semantic domains, and fine-grained visual concepts.
    \item We introduce \expertlens, a data-free method that decodes router weights to identify the semantic specialization of individual experts.
    \item We demonstrate that emergent modularity enables efficient multimodal adaptation, matching full fine-tuning and outperforming LoRA while reducing training time by over $4\times$ on average.
\end{itemize}

%% file: sections/related.tex
\section{Related Work}

\mypar{Mixture-of-Experts.}
Mixture-of-Experts (MoE) architectures have a long history in conditional computation, scaling model capacity by sparsely routing each input through a subset of expert modules~\citep{shazeer2017, jacobs1991adaptive}. Sparse MoEs were later incorporated into large-scale Transformers~\citep{lepikhin2020gshard, fedus2022switch, jiang2024mixtral, dai_deepseekmoe_2024}, and are now common for scaling modern language models~\citep{team2026kimi, xu2026deepseek} and multimodal foundation models~\citep{qwen3vl, gemma4, kimivl, inkling}. We provide a brief overview of MoE routing and notation in \S\ref{subsec:moes}.

\mypar{Interpreting Internal Representations.} Mechanistic interpretability seeks to reverse-engineer neural networks into human-understandable mechanisms using tools such as circuit analysis, causal interventions, and representation decoding~\citep{elhage2021mathematical, meng_locating_2023, conmy2023automatedcircuitdiscoverymechanistic}. Logit Lens~\citep{nostalgebraist2020logitlens} applies a model's output head to intermediate residual-stream representations to expose how predictions evolve across layers. This is motivated by the residual stream's approximately linear, additive structure~\citep{elhage2021mathematical} and evidence that transformer MLPs write semantically meaningful, vocabulary-aligned updates into it~\citep{geva2022transformerfeedforwardlayersbuild, wu2025semantichubhypothesislanguage}. Subsequent variants have extended this idea to decoding internal representations in LLMs and VLMs~\citep{belrose2025elicitinglatentpredictionstransformers, Pal_2023, neo2025interpretingvisualinformationprocessing}.

Interpretability of Mixture-of-Experts (MoE) models remains comparatively underexplored. Prior work has used attribution and ablation to identify behaviorally important experts~\citep{hu_what_2026, li_decoding_2025}, manipulated routing or expert activation to steer behavior~\citep{fayyaz_steering_2025}, and decoded individual expert contributions into output space~\citep{chaudhari_moe_2026}. We extend this line of work with \expertlens, a data-free method for decoding expert semantics and identifying domain-specialized experts for efficient adaptation.

\mypar{Expert Specialization.} Prior work on text-only MoEs has found emergent expert structure, including specialization by token type and part of speech~\citep{jiang2024mixtral, antoine2025part, li_understanding_2026}. However, this structure is typically low-level rather than aligned with broader semantic domains. To encourage higher-level organization, EMO~\citep{wang2026emo} modifies text-only MoE pretraining to promote domain-level modularity. Expert organization in multimodal MoEs remains less explored, with recent work examining routing across visual, textual, and reasoning stages~\citep{xu2026seeing, zhang2026beyond}. In contrast, we find that pretrained multimodal MoEs develop rich semantic organization spanning modalities, domains, and fine-grained visual concepts.

\mypar{Parameter Efficient Adaptation.} Parameter-efficient fine-tuning methods such as LoRA reduce adaptation cost by introducing a small number of trainable parameters while keeping the pretrained model frozen~\citep{hu2022lora}. Recent work extends low-rank adaptation to MoE models, applying LoRA updates to shared and/or expert parameters~\citep{wei2026hellora, liu2026parameter}. In contrast, our approach exploits the pretrained model's existing expert structure, selectively updating the experts most relevant to the target domain. In direct comparison, \expertlens achieves higher adaptation performance than LoRA while also reducing training time more substantially.

%% file: sections/em_structure.tex
\section{Emergent Structure in  Mixture-of-Experts VLMs}
\label{sec:emergent_structure}
Mixture-of-Experts models provide a natural opportunity for functional specialization within their expert modules, but it remains unclear whether such specialization emerges in pretrained models and, if so, how it is organized. We explore this idea via three questions:

\begin{enumerate}[nosep]
\item Do experts specialize on different modalities?  
\item Do experts specialize on different semantic domains within a modality? 
\item Does this specialization persist at finer-grained, object-level semantics? 
\end{enumerate}

Answering these questions can provide insight into how multimodal MoEs organize their internal representations, while also revealing whether their modular structure can be exploited for efficient, targeted adaptation. We study these questions across three frontier multimodal MoEs: \qwen~\citep{qwen3vl}, \kimi~\citep{kimivl}, and \gemma~\citep{gemma4}. We conclude by introducing \expertlens, a data-free method for identifying domain-specialized experts from expert router weights.

Throughout this section, we use a suite of VQA datasets spanning general, medical, math/scientific, and remote-sensing domains to probe MoE expert behavior across image patches and text tokens. For text-only analyses, we additionally include broader linguistic domains such as legal and abstract reasoning. We include all datasets used and additional experimental details in \S\ref{appx:visualization_details}.

\subsection{Expert Specialization by Modality}
\label{sec:expertspecialization_modality}

\begin{figure}
    \centering
    \includegraphics[width=0.98\linewidth]{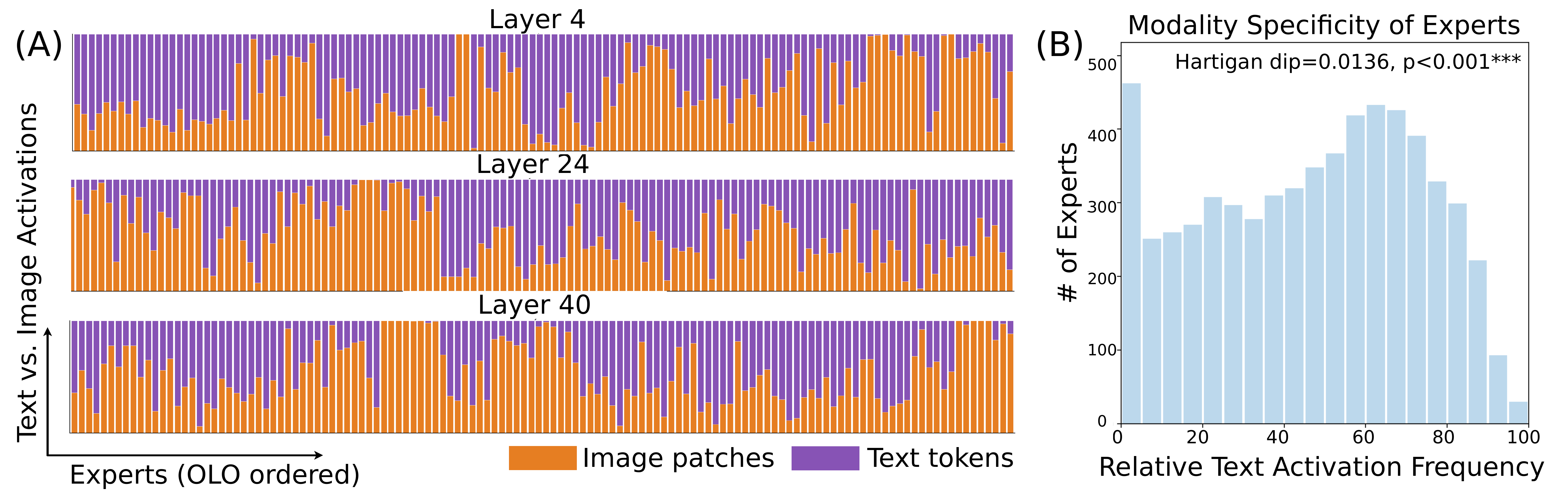}
    \vspace{-1mm}
\caption{\textbf{Experts specialize by modality.}
\textbf{(A)}: Fraction of each expert's top-$K$ $(K=8)$ activations from image patches versus text tokens, ordered by optimal leaf ordering (OLO; \S\ref{appx:olo}). Most experts strongly favor one modality. \textbf{(B)}: Distribution of each expert's text-token activation share. For \qwen{}, the distribution is bimodal (Hartigan dip $=0.0136$, $p<0.001$).}
    \label{fig:image_text_ratio}
    \vspace{-1mm}
\end{figure}

% , with $433$ text-leaning experts at $60$--$65\%$ and $462$ image-exclusive experts at $0$--$5\%$.

We first ask whether multimodal MoEs exhibit modality-specific routing. In particular, do individual experts preferentially process visual or textual tokens, and is specialization distributed similarly across modalities? Establishing such structure would indicate that experts in pretrained multimodal MoEs are not interchangeable, but develop distinct functional preferences. To this end, we measure modality affinity from each expert's activation counts on modality-specific tokens. We sample $3{,}840$ image patches and equally many text tokens from seven datasets spanning general VQA, medical imaging, and math/science reasoning. Under the expert activation definition (\S\ref{subsec:moes}), we define affinity as the fraction of each expert's top-$K$ activations from image versus text tokens.

\mypar{Experts develop modality-specific routing.}
Figure~\ref{fig:image_text_ratio}A shows, for each expert, the fraction of top-$K$ activations arising from image versus text tokens. The x axis indexes experts within each layer; orange-dominated experts favor image tokens, and purple-dominated experts favor text. Across layers, activations are rarely balanced between modalities. Instead, many experts strongly prefer one, indicating that modality specialization emerges naturally in pretrained multimodal MoEs.

\mypar{Modality specialization tends to be bimodal and asymmetric.}
We pool experts across layers by text-token activation share. Without modality preference, the distribution should peak near $50\%$; instead, \qwen is clearly bimodal (Figure~\ref{fig:image_text_ratio}B), with an image-leaning mode at $0$--$5\%$ text share ($462$ experts) and a text-leaning mode at $60$--$65\%$ ($433$ experts). Hartigan's dip test rejects unimodality (dip $=0.0136$, $p<0.001$). Image specialization is more extreme, while text specialization spreads across $60$--$80\%$. \gemma is similarly bimodal, while \kimi is more nearly unimodal, likely due to its smaller expert pool (\S\ref{appx:modality}).

\subsection{Domain Specificity in Vision and Text Experts}
\label{subsec:domain_specificity}

Having established modality preferences, we next ask whether specialization extends to semantic domains. Do some experts preferentially process medical over math inputs? We test whether visual and textual representations organize differently by domain, and whether this structure reflects semantics rather than linguistic factors or low-level visual statistics. We first examine this with t-SNE projections of token activations and router weights, then quantify it using kNN purity.

\begin{figure}
    \centering
    \includegraphics[width=0.9\linewidth]{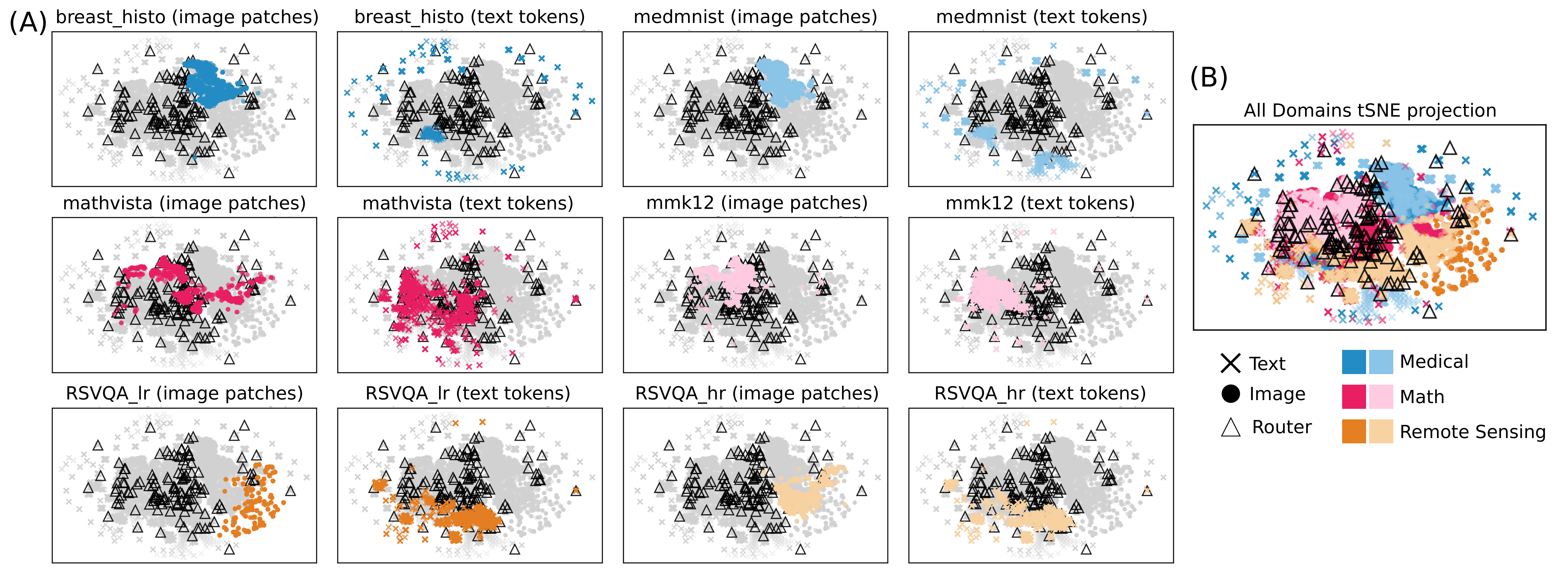}
    \vspace{-1mm}
    \caption{\textbf{Visual activations exhibit stronger domain structure than textual activations.}
We project layer-18 text-token and image-patch activations with router weights from \qwen using t-SNE. \textbf{(A)}: Datasets are colored and separated by modality; image activations form tighter domain-specific clusters than text. \textbf{(B)}: All domains shown together.}
    \vspace{-2mm}
    \label{fig:tsne_isolated_domains}
\end{figure}

\mypar{Visual activations exhibit strong domain structure.}
We ask whether modality specialization extends to semantic domains by examining the representations that drive expert routing. Because routing depends directly on token activations, domain structure in this space could induce domain-specific expert use. We sample $1.5$K image patches and text tokens from each of six VQA datasets spanning medical, math, and remote sensing domains ($18$K activations total), and jointly project them with all the router weights in \qwen using t-SNE. Figure~\ref{fig:tsne_isolated_domains} shows layer 18 as an exemplar, with similar patterns across layers and models (\S\ref{appx:tsne}).

Figure~\ref{fig:tsne_isolated_domains} reveals two trends. First, image-patch activations form tighter domain-specific clusters than text-token activations, with datasets from the same domain overlapping substantially. For example, breast histopathology and \medmnist cluster together, as do \rsvqalr and \rsvqahr, suggesting that the structure reflects domain rather than dataset identity. Second, router weights concentrate near image activations, while text activations are more diffuse. Together, these observations suggest that visual representations exhibit the domain structure that could induce domain-specific routing.

\begin{figure}
    \centering
    \includegraphics[width=1.0\linewidth]{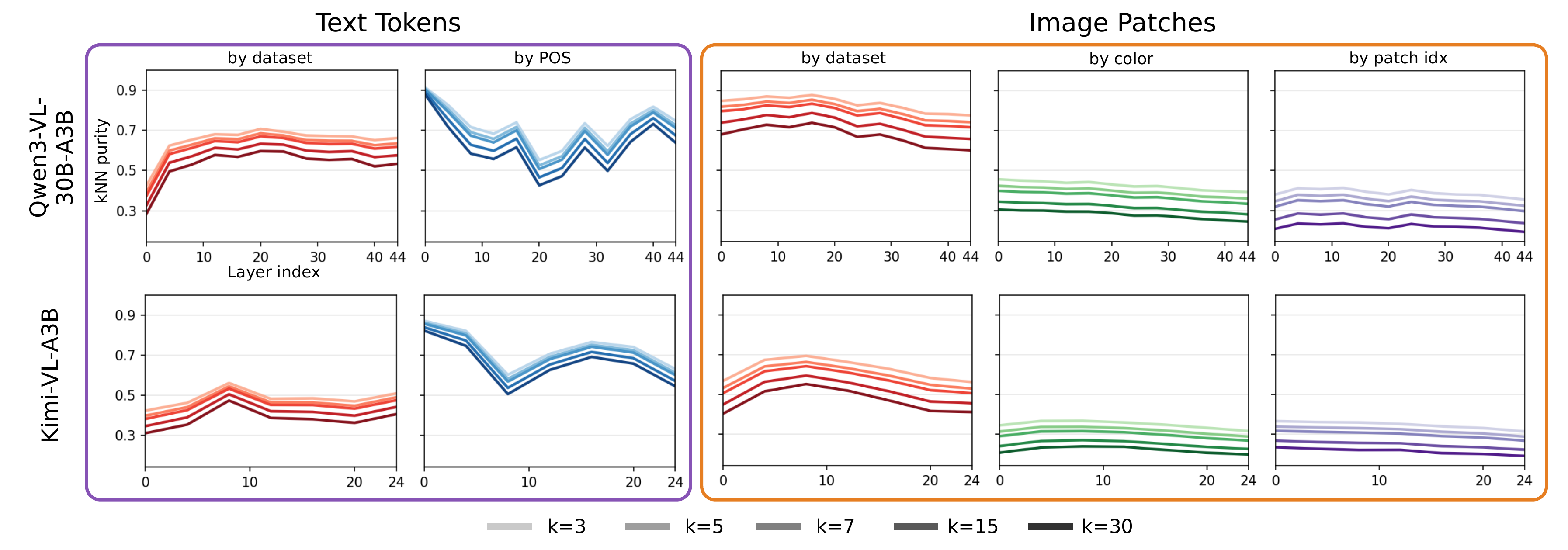}
        \vspace{-7mm}
        \caption{\textbf{Visual activations organize by domain; textual activations by linguistic structure.}
        We measure kNN purity across layers, grouping text activations by dataset or part of speech and image activations by dataset, color, or spatial position. Text activations align most with part of speech, while image activations align best with semantic domain.}
        \vspace{-4mm}
    \label{fig:knn_purity}
\end{figure}

\mypar{Quantifying domain structure with kNN purity.}
The t-SNE analysis suggests strong domain structure in visual activations, but not which factors best explain it. We therefore quantify local activation-space structure using kNN purity: for each activation, we measure the fraction of its $k$ nearest neighbors sharing the same label, where higher purity indicates better alignment with the local representation geometry. For text tokens, we compare groupings by dataset or part of speech; for image patches, by dataset, mean color, or spatial position. Each grouping contains 12 balanced categories. Further details are in \S\ref{appendix:knn}. We report results across several values of $k$ in Figure~\ref{fig:knn_purity}.

\mypar{Vision and text representations organize differently.}
Figure~\ref{fig:knn_purity} shows that text-token activations are better organized by part of speech than by domain, consistent with the findings in prior analyses of the experts of text-only MoEs~\citep{antoine2025part,jiang2024mixtral}. In contrast, image-patch activations achieve highest kNN purity when grouped by dataset, substantially exceeding mean color or spatial patch position. This pattern holds across layers and both \qwen and \kimi. These results support the t-SNE analysis and suggest that visual domain organization is not explained by low-level properties such as color or spatial location.

\subsection{Visual Expert Specialization by Object Semantics}
\label{subsec:semantic_specialization}

With domain structure in visual representations established, we then ask if specialization extends to finer-grained semantics. We test whether visually distinct but semantically related classes recruit similar expert sets. To isolate semantics from dataset effects, we conduct the analysis entirely within TinyImageNet~\citep{le2015tiny}, comparing overlap between related and unrelated classes.

\mypar{Measuring semantic expert overlap.}
We use all 200 TinyImageNet classes, forming $19{,}900$ object-class pairs spanning a range of semantic similarity, and sample 100 images per class. For each image, we track which experts activate on its patch tokens. An expert counts as active only if it activates on at least $m$ patches, for $m\in\{3,5,8,12,16\}$. For each class and layer, we retain experts active in at least $60\%$ of that class's images. We then measure Jaccard overlap between each class pair's expert sets and standardize it with a score $Z(a,b)$ relative to all class pairs, where larger values indicate stronger shared specialization. Details and score calculations are in \S\ref{appendix:tinyimagenet}.

\begin{figure}
    \centering
\includegraphics[width=1.0\linewidth]{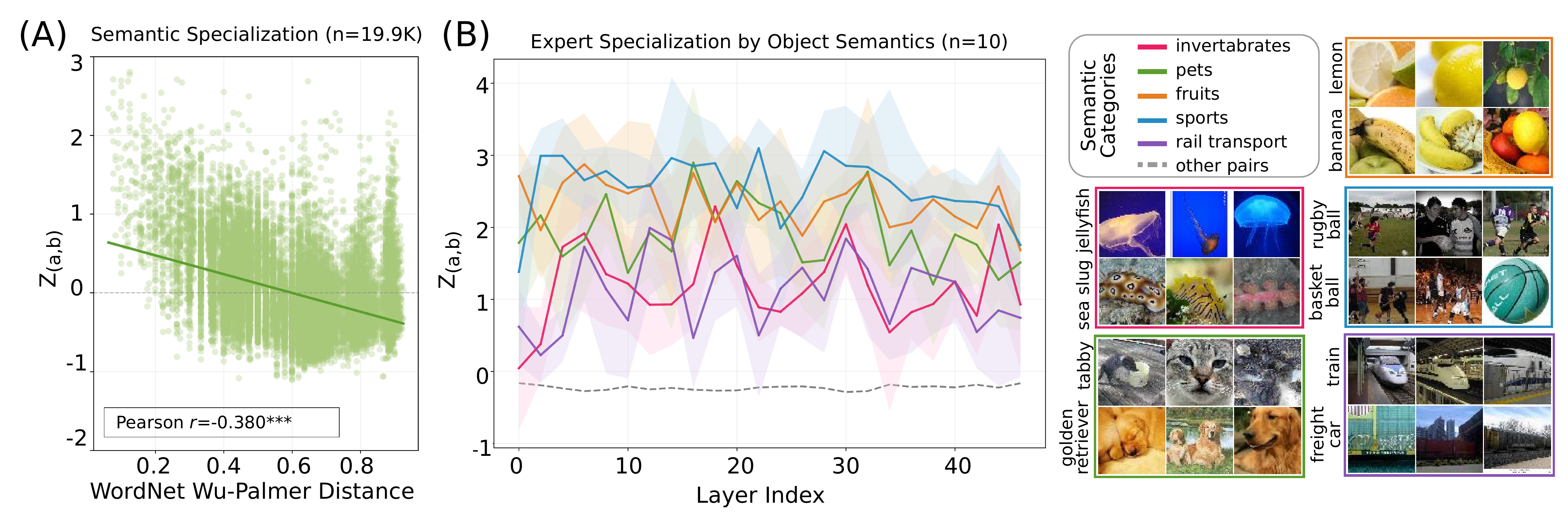}
    \vspace{-7mm}
    \caption{\textbf{Semantically related object classes recruit similar experts.}
\textbf{(A)} Across all TinyImageNet class pairs, expert-set overlap $Z(a,b)$ decreases with semantic distance (Pearson $r=-0.380$, $p<0.001$). \textbf{(B)} For 10 classes forming five semantic pairs, related classes show greater expert overlap across layers despite visual variation. Solid lines average thresholds $m\in\{3,5,8,12,16\}$, shading denotes $\pm1$ standard deviation, and the gray dashed line shows unrelated pairs.}
    \label{fig:semantic_tinyimagenet}
    \vspace{-3mm}
\end{figure}

\mypar{Semantically related classes activate similar experts.}
Figure~\ref{fig:semantic_tinyimagenet} shows that related class pairs share more experts than unrelated pairs across layers. In Figure~\ref{fig:semantic_tinyimagenet}A, semantic distance measured by WordNet Wu-Palmer distance~\citep{wu-palmer-1994-verb} is inversely correlated with expert-set overlap: larger distance indicates lower semantic relatedness. For the purpose of providing an overview, the semantic overlap scores are shown averaged across all activation thresholds and all model layers. Figure~\ref{fig:semantic_tinyimagenet}B provides a finer-grained view using 10 TinyImageNet classes forming five semantically related pairs, without averaging across layers or patch thresholds. Their specialization scores remain above the gray unrelated-class baseline, with several pairs reaching $Z(a,b)>3$ in intermediate layers, i.e., more than three standard deviations above baseline. These results show that visual expert specialization extends beyond coarse domains to finer-grained object semantics.

\mypar{Semantic overlap persists despite visual variation.}
Importantly, high expert overlap is not limited to visually similar classes. Figure~\ref{fig:semantic_tinyimagenet}B shows that while some related pairs, such as trains and freight cars, share visual structure, others, such as jellyfish and sea slugs or tabbies and golden retrievers, differ substantially in appearance and background. Yet they recruit similar expert sets. We observe the same object-level specialization in other MoE-VLMs (\S\ref{appx:addtl_semantic_vis}). Together, our analyses suggest that visual expert specialization reflects semantics rather than dataset identity or low-level similarity.

\begin{figure}
    \centering
\includegraphics[width=0.97\linewidth]{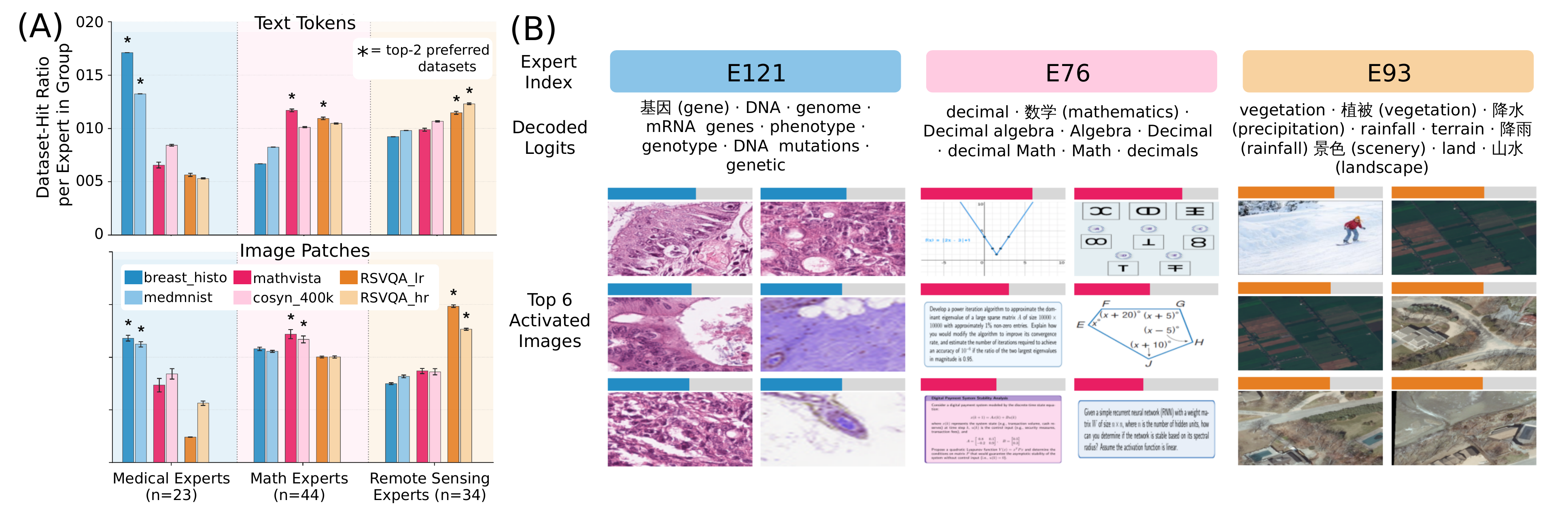}
    \vspace{-3mm}
    \caption{\textbf{\expertlens selects domain-specific experts.}
\textbf{(A)} \expertlens-derived cohorts preferentially activate on their target domains. Bar colors denote datasets; heights show normalized cohort firing. Across most cohorts, activation is strongest on the corresponding domain. \textbf{(B)} Representative experts with their decoded logits and highest-activating images: E121 is medical, E76 math, and E93 remote sensing. All results use layer 18 of \qwen.}
    \label{fig:expertlens_viewing_ab}
    \vspace{-5.5mm}
\end{figure}

\subsection{ExpertLens: Identifying Domain-Specialized Experts}
\label{subsec:expertlens}

Our analyses show that multimodal MoEs develop domain specialization, but identifying these experts has required domain-specific examples to observe routing behavior. This is limiting for targeted adaptation, where we would prefer to identify experts for a new domain without representative data. We therefore ask: \emph{can we recover an expert's semantic specialization from its routing weights?} We introduce \expertlens, a data-free method that decodes semantic preferences from router weights.

\mypar{Introducing ExpertLens.}
\expertlens builds on a simple property of MoE routing: an expert is selected when its router weight strongly aligns with a token representation. If token representations encode semantics, then a domain-specialized expert's router weight should align with semantically related directions. We thus map each router weight through the model's LM head to obtain vocabulary-token logits, analogous to logit-lens decoding~\citep{belrose2025elicitinglatentpredictionstransformers}. For expert $i$ in layer $\ell$, let $w_i^{(\ell)}$ denote its router weight and $U$ the LM head. We compute:

\vspace{-5mm}
\begin{equation}
    z_i^{(\ell)} = U w_i^{(\ell)},
\qquad
\mathcal{V}_M^{(\ell,i)}
=
\operatorname{TopM}\!\left(z_i^{(\ell)}\right),
\qquad
S_{\mathrm{EL}}^{(\ell,i)}(d)
=
\frac{1}{M}
\sum_{v \in \mathcal{V}_M^{(\ell,i)}}
\mathrm{Rel}(v,d).
\end{equation}
\vspace{-5mm}

where $\mathcal{V}_M^{(\ell,i)}$ is the set of top-$M$ tokens decoded from expert $i$ in layer $\ell$, and $\mathrm{Rel}(v,d)\in\{0,1\}$ indicates whether token $v$ is semantically related to domain $d$, as judged by the model (details in \S\ref{appx:el_classification}). The \expertlens score is thus the fraction of an expert's top-$M$ decoded tokens associated with the target domain. In all experiments, we use $M=10$ and select experts with $S_{\mathrm{EL}}\geq0.9$.

\mypar{Evaluating ExpertLens.}
We test whether \expertlens specializations predict expert use on domain-specific inputs. Fig.~\ref{fig:expertlens_viewing_ab}A shows layer-18 routing of \expertlens-identified medical ($n{=}23$), math ($n{=}44$), and remote sensing ($n{=}34$) experts in \qwen over text and image tokens from six VQA datasets. For each expert, we compute its activation fraction by dataset and normalize by cohort size. This gives experts equal weight, preventing cohort size or firing rate from dominating. Error bars show SEM; asterisks mark the two highest-scoring datasets.

\mypar{ExpertLens uncovers domain specificity.}
Fig.~\ref{fig:expertlens_viewing_ab}A shows that datasets preferentially recruit experts whose \expertlens labels match their domain. Medical, math, and remote-sensing datasets most strongly recruit their corresponding expert cohorts, except for weaker math alignment on text tokens. Thus, experts identified without domain-specific data still align with the domains that preferentially recruit them. Although Fig.~\ref{fig:expertlens_viewing_ab}A shows layer 18, the same pattern holds across layers (\S\ref{appx:el_classification}). Fig.~\ref{fig:expertlens_viewing_ab}B shows this correspondence at the level of individual experts. For representative medical (E121), math (E76), and remote sensing (E93) experts, the highest-logit decoded tokens align semantically with the domain of the images that most strongly recruit each expert. This indicates that \expertlens's decoded preferences closely track the inputs that activate each expert.

%% file: sections/experiments.tex
\section{Domain-Specialized Experts for Efficient Adaptation}
\label{sec:experiments}
In \S\ref{sec:emergent_structure} we observed that multimodal MoEs develop domain-specialized experts. We now ask whether this structure can enable more efficient adaptation. In particular, can selectively adapting domain-relevant experts match full fine-tuning while reducing post-training parameters and compute?

\mypar{Sequential Adaptation Protocol.}
To study this, we adapt a single model sequentially to math, medical, and remote sensing. Rather than resetting to the pretrained checkpoint between domains, this protocol tests whether targeted expert adaptation remains effective as the same model is repeatedly adapted. We evaluate all three domains throughout training and study domain ordering in \S\ref{subsec:ablations_analysis}.

\mypar{Selective Adaptation.}
We identify domain experts with the \expertlens method (\S\ref{subsec:expertlens}) and select those with $S_{\mathrm{EL}}\geq0.9$. For \qwen, this yields $2{,}768$, $1{,}103$, and $1{,}290$ experts for math, medical, and remote sensing. The sets are largely distinct (Jaccard overlap: $0.13\%$ math--medical, $6.48\%$ math--remote sensing, $13.20\%$ medical--remote sensing). We tune selected expert MLPs, freezing all others and the vision encoder. We also train shared decoder parameters, including QKV projections, router weights, and the LM head (\S\ref{subsec:ablations_analysis}). As expert MLPs contain most model parameters, $53.0$--$78.3\%$ of the model is frozen. We call this procedure \expertlens throughout.

\mypar{Train and Evaluation Setup.}
We train on $4{,}000$ examples per domain and evaluate on multiple benchmarks spanning math, medical, and remote sensing. Training datasets, evaluation benchmarks, and implementation details are in \S\ref{appx:experiment_details}. We use the system prompt and rewards from VERO~\citep{vero}, \qwen as the base MoE, and GSPO~\citep{gspo} for post-training. We train for $200$ steps per domain on $4$ H200 GPUs with batch size $32$ and group size $5$. Results use three seeds, with shaded regions showing $95\%$ confidence intervals.

\begin{figure}[t]
\centering
\includegraphics[width=\linewidth]{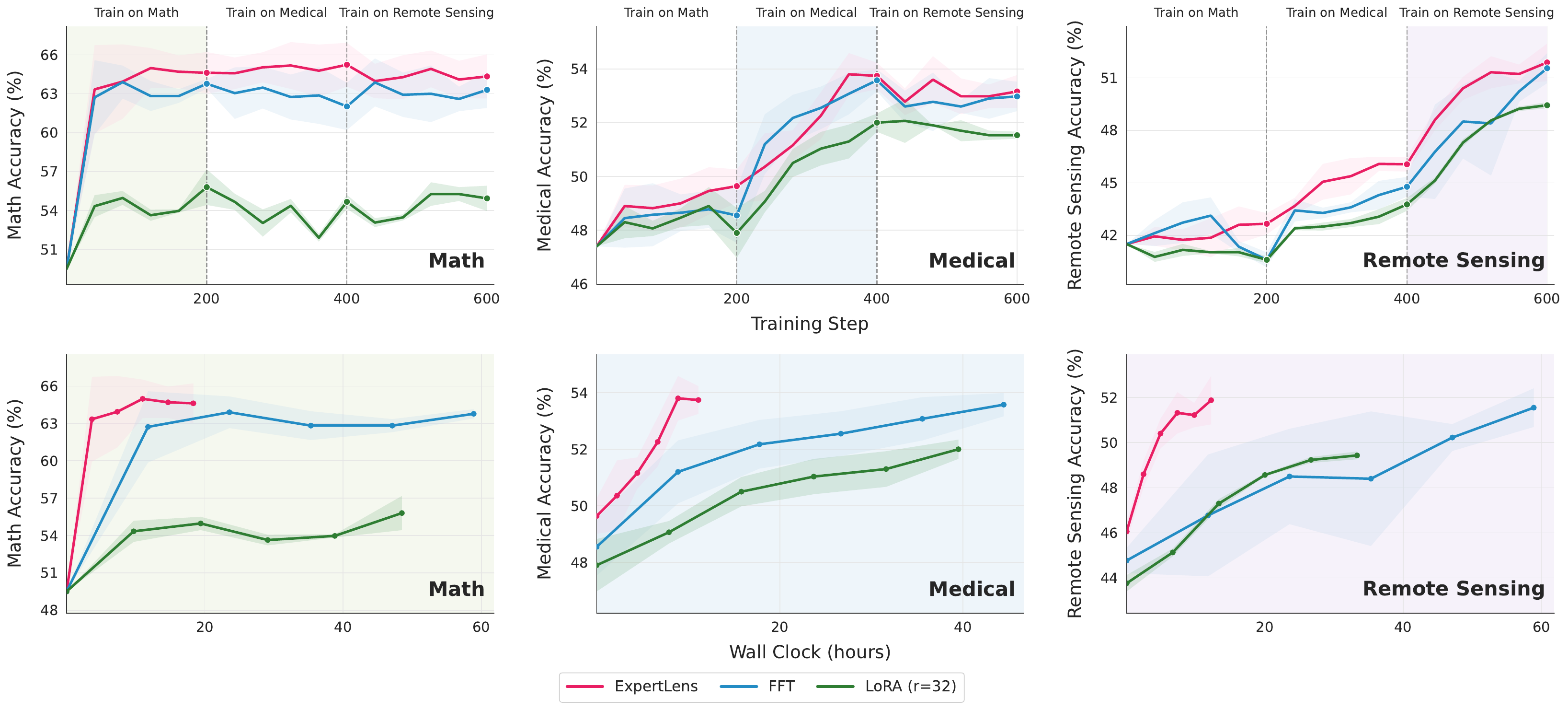}
\vspace{-7mm}
\caption{\textbf{ExpertLens matches FFT, outperforms LoRA, and accelerates adaptation.}
\textbf{(A)} \expertlens matches or surpasses FFT while outperforming LoRA. \textbf{(B)} Highlighted regions show the same stages as \textbf{(A)}, shown against wall-clock time. \expertlens achieves a $4.0\times$ average speedup over FFT while matching performance, and a $2.9\times$ speedup over LoRA.}
\vspace{-4mm}
\label{fig:el_fft_trajectories}
\end{figure}

\subsection{ExpertLens for Efficient Multimodal Adaptation}
\label{subsec:el_gspo}

We evaluate whether selectively fine-tuning domain-specialized experts provides an effective alternative to both full-model and parameter-efficient adaptation. We compare \expertlens against full fine-tuning (FFT) and LoRA~\citep{hu2022lora} with the same data, objective, and sequential adaptation protocol. For LoRA, we evaluate ranks $R\!\in\!\{32,64\}$ with $\alpha\!=\!2R$ following prior work~\citep{shuttleworth2025lora}, and report the best configuration ($R\!=\!32$).

\mypar{ExpertLens matches FFT and outperforms LoRA.}
Figure~\ref{fig:el_fft_trajectories}A shows \expertlens matches or surpasses FFT throughout adaptation while updating only $21.7$--$47.0\%$ of model parameters. At the end of each training stage, \expertlens achieves $64.6\%$ versus $63.8\%$ on math, $53.7\%$ versus $53.6\%$ on medical, and $51.9\%$ versus $51.6\%$ on remote sensing. LoRA reaches $55.8\%$, $52.0\%$, and $49.4\%$, respectively, below both methods. \expertlens thus identifies sufficient domain-specialized capacity to match full-model adaptation and outperforms LoRA.

\mypar{ExpertLens accelerates adaptation.}
Selective adaptation also substantially reduces training time. Figure~\ref{fig:el_fft_trajectories}B shows in-domain performance against wall-clock time. Relative to FFT, \expertlens achieves speedups of $3.2\times$ on math, $4.0\times$ on medical, and $4.8\times$ on remote sensing, averaging $4.0\times$ over the entire training sequence. Compared with LoRA, \expertlens trains $2.9\times$ faster and achieves higher performance across all domains. Thus, exploiting MoE modularity can yield better performance--efficiency trade-off than FFT or LoRA.

\subsection{The Importance of Expert Selection}
\label{subsec:expert_selection}
\begin{figure}[t]
\centering
\includegraphics[width=0.93\linewidth]{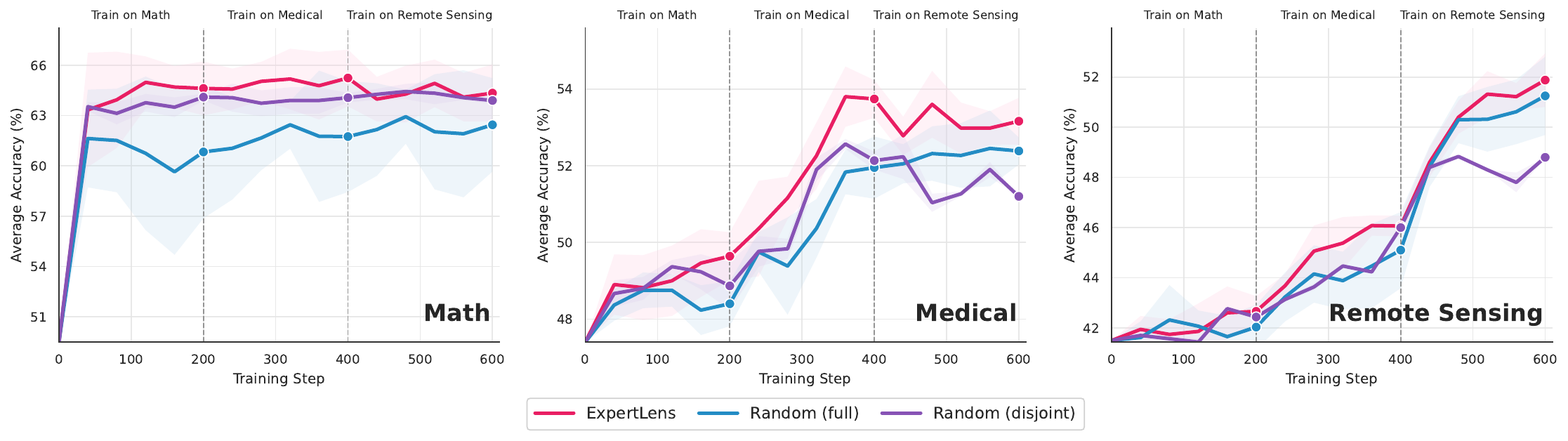}
\vspace{-3.5mm}
\caption{\textbf{ExpertLens outperforms matched-budget random expert selection.}
\expertlens outperforms both matched-budget random selection baselines while updating the same number of experts.}
\vspace{-5mm}
\label{fig:el_random}
\end{figure}

We next ask whether \expertlens gains arise from sparsity alone or from the identity of the experts being trained. In Figure~\ref{fig:el_random}, we compare against two random selection baselines with the same expert budget per domain as \expertlens. \emph{Random (full)} samples from the full expert pool, while \emph{Random (disjoint)} samples only experts not selected by \expertlens. Expert subsets are resampled for each seed for both baselines. This comparison isolates the benefit of domain-relevant expert selection beyond simply updating fewer parameters.

\mypar{Sparsity alone does not explain the gains.}
We first compare \expertlens with \emph{Random (full)}, which updates the same number of experts sampled from the full expert pool. Figure~\ref{fig:el_random} shows \expertlens consistently outperforms \emph{Random (full)}, with the clearest gains on math and medical. At the end of each training stage, \expertlens achieves $64.6\%$ versus $60.8\%$ on math, $53.7\%$ versus $52.0\%$ on medical, and $51.9\%$ versus $51.3\%$ on remote sensing. Since both methods update the same number of experts, sparsity alone cannot explain the gains; expert identity matters.

\mypar{ExpertLens-selected experts are particularly effective.}
We next compare against \emph{Random (disjoint)}, which samples the same number of experts only from those not selected by \expertlens. Figure~\ref{fig:el_random} shows that this baseline matches \expertlens on math but falls behind on medical and remote sensing. Thus, while alternative expert subsets can suffice for some domains, \expertlens-selected experts are more effective overall. Together with \emph{Random (full)}, this shows that selective fine-tuning depends not only on sparsity, but on selecting domain-relevant experts.

\input{tables/naive_oracle_comp}
\mypar{ExpertLens approaches privileged selection.}
We compare \expertlens with two count-based selectors with the same expert budget. \textit{Frequency Selection} ranks experts by current-domain activation counts, while the \textit{Privileged Oracle} ranks by current-domain counts minus counts from other domains. The oracle requires data from all domains, including future ones, and depends on the datasets used. Despite using no domain examples, ExpertLens achieves comparable final accuracy to both baselines (Table~\ref{table:naive_oracle_per_domain}). Thus, \expertlens matches count-based selection while remaining usable without target examples or future-domain knowledge.

\subsection{Ablations and Analyses}
\label{subsec:ablations_analysis}

We next examine the design choices underlying \expertlens and the robustness of selective expert adaptation. We first ablate which parameter groups are updated alongside the selected experts, then evaluate whether our conclusions depend on the ordering of the training domains. 
\input{tables/freezing_ablations}

\mypar{Shared decoder parameters are important for effective adaptation.}
Table~\ref{table:freezing_ablations} ablates which parameters are trained alongside \expertlens-selected experts. \emph{In-domain} performance is accuracy after training on a domain, while \emph{out-of-domain} performance averages the other two. Training only selected expert MLPs yields weaker adaptation, especially on math. Adding the router improves math by $11.2$ points, with QKV projections and the language-model head providing further gains across domains. These results suggest that selective adaptation benefits from updating the shared parameters that route inputs to the trained experts and shape their representations.

\mypar{ExpertLens is robust to domain ordering.}
To evaluate whether our findings are sensitive to domain ordering, we repeat adaptation under three training orders in Table~\ref{table:domain_ordering_per_domain}. \expertlens outperforms FFT in mean final accuracy under all three, by $1.7$, $2.0$, and $0.2$ percentage points, respectively. It is also less sensitive to ordering, with the standard deviation across orders decreasing from $1.37$ to $0.78$ percentage points ($43\%$). Although gains vary across orders, the overall benefit of selective adaptation persists across different training sequences.

\input{tables/domain_ordering}

\subsection{Experiments on Kimi-VL}
\label{subsec:kimi_exps}

\input{tables/kimi_vl}
To test whether targeted adaptation generalizes beyond \qwen, we repeat our main experiment on \kimi, comparing \expertlens with full fine-tuning and matched-budget \emph{Random (full)}. Using the same threshold, $S_{\mathrm{EL}}\geq0.9$, selects $560$, $301$, and $349$ experts for math, medical, and remote sensing, respectively. For the semantic relevance function $\mathrm{Rel}(\cdot)$, we use \qwen to classify decoded tokens, as \kimi was less reliable at this task.

\mypar{ExpertLens maintains its advantage on Kimi-VL.}
Table~\ref{table:kimi_vl_final_per_domain} reports final accuracy after the sequential adaptation sequence. \expertlens achieves the highest mean performance at $48.0\%$, compared with $47.4\%$ for FFT and $45.0\%$ for random selection, showing that targeted expert adaptation extends beyond \qwen. \expertlens also remains faster to train than FFT on \kimi, providing a $1.53\times$ speedup over the full sequence.

%% file: tables/naive_oracle_comp.tex
\setlength{\columnsep}{8pt}

\begin{wraptable}{r}{0.52\textwidth}
\vspace{-4mm}
\small

\resizebox{\linewidth}{!}{%
\begin{tabular}{@{}l|ccc|c@{}}
\toprule
Selection Method
& Math
& Medical
& Remote Sensing
& Mean \\
\midrule

Frequency Selection
  & 63.8
  & 52.7
  & 51.6
  & 56.0 \\

Privileged Oracle
  & 63.0
  & \textbf{55.5}
  & \textbf{52.1}
  & \textbf{56.9} \\

\expertlens
  & \textbf{64.3}
  & 53.2
  & 51.9
  & 56.5 \\

\bottomrule
\end{tabular}%
}
\vspace{-2mm}
\caption{\textbf{ExpertLens approaches privileged expert selection without using domain data for expert identification.}}
\label{table:naive_oracle_per_domain}
\vspace{-3mm}
\end{wraptable}

%% file: tables/freezing_ablations.tex
\begin{table}[t]
\centering
\resizebox{0.99\linewidth}{!}{
\begin{tabular}{l *{6}{r@{\;}l}}
\toprule
& \multicolumn{6}{c}{In-domain}
& \multicolumn{6}{c}{Out-of-domain} \\
\cmidrule(lr){2-7}\cmidrule(lr){8-13}
Trainable Parameters
& \multicolumn{2}{c}{Math}
& \multicolumn{2}{c}{Medical}
& \multicolumn{2}{c}{Remote Sensing}
& \multicolumn{2}{c}{Math}
& \multicolumn{2}{c}{Medical}
& \multicolumn{2}{c}{Remote Sensing} \\
\midrule

Selected Experts Only
& 52.2 & 
& 49.4 & 
& 48.7 & 
& 44.7 & 
& 47.2 & 
& 51.7 & \\

+ Router Weights
& 63.4 & \gain{+11.2}
& 50.7 & \gain{+1.3}
& 49.3 & \gain{+0.6}
& 45.7 & \gain{+1.0}
& 54.6 & \gain{+7.4}
& 56.2 & \gain{+4.5} \\

+ QKV Projections
& 65.8 & \gain{+2.4}
& 53.1 & \gain{+2.4}
& 52.3 & \gain{+3.0}
& 45.5 & \gain{-0.2}
& 55.0 & \gain{+0.4}
& 57.7 & \gain{+1.5} \\

+ LM Head (\expertlens)
& \textbf{66.1} & \gain{+0.3}
& \textbf{54.1} & \gain{+1.0}
& \textbf{52.8} & \gain{+0.5}
& \textbf{46.1} & \gain{+0.6}
& \textbf{56.1} & \gain{+1.1}
& \textbf{59.7} & \gain{+2.0} \\
\bottomrule
\end{tabular}
}
\vspace{-2mm}
\caption{\textbf{Shared decoder parameters are important for effective adaptation.}
Progressively training the router, attention weights, and language-model head alongside the selected experts improves performance on both in-domain and out-of-domain samples, with the full \expertlens configuration performing best across domains.}
\label{table:freezing_ablations}
% \vspace{-2mm}
\end{table}

%% file: tables/domain_ordering.tex
\begin{table}[t]
\centering
\small
\begin{tabular}{
  @{}
  ll
  c@{\hspace{3.4em}}
  c@{\hspace{3.4em}}
  c@{\hspace{3.4em}}
  |
  c@{\hspace{3.2em}}
  @{}
}
\toprule
Domain Ordering
& Method
& Math
& Medical
& Remote Sensing
& Mean \\
\midrule

\multirow{2}{*}{
  Math {\scriptsize$\rightarrow$}
  Med.\ {\scriptsize$\rightarrow$}
  Rem.\ Sens.
}
& FFT
  & 62.4
  & 52.6
  & 51.0
  & 55.3 \\
& \expertlens
  & \textbf{65.2}\rlap{\;\scriptsize\gain{+2.8}}
  & \textbf{53.2}\rlap{\;\scriptsize\gain{+0.6}}
  & \textbf{52.6}\rlap{\;\scriptsize\gain{+1.6}}
  & \textbf{57.0}\rlap{\;\scriptsize\gain{+1.7}} \\

\cmidrule(lr){1-6}

\multirow{2}{*}{
  Med.\ {\scriptsize$\rightarrow$}
  Math {\scriptsize$\rightarrow$}
  Rem.\ Sens.
}
& FFT
  & 57.1
  & 51.5
  & 51.9
  & 53.5 \\
& \expertlens
  & \textbf{61.3}\rlap{\;\scriptsize\gain{+4.2}}
  & \textbf{52.3}\rlap{\;\scriptsize\gain{+0.8}}
  & \textbf{52.7}\rlap{\;\scriptsize\gain{+0.8}}
  & \textbf{55.5}\rlap{\;\scriptsize\gain{+2.0}} \\

\cmidrule(lr){1-6}

\multirow{2}{*}{
  Rem.\ Sens.\ {\scriptsize$\rightarrow$}
  Med.\ {\scriptsize$\rightarrow$}
  Math
}
& FFT
  & 62.6
  & \textbf{52.9}
  & \textbf{53.1}
  & 56.2 \\
& \expertlens
  & \textbf{67.2}\rlap{\;\scriptsize\gain{+4.6}}
  & 51.0\rlap{\;\scriptsize\textcolor{red}{(-1.9)}}
  & 50.9\rlap{\;\scriptsize\textcolor{red}{(-2.2)}}
  & \textbf{56.4}\rlap{\;\scriptsize\gain{+0.2}} \\

\bottomrule
\end{tabular}
\vspace{-2mm}
\caption{\textbf{ExpertLens is robust to domain ordering.}
\expertlens achieves higher mean performance than full fine-tuning across all three domain orderings, with less variation across sequences. Parentheses denote the difference relative to FFT.}
\vspace{-2mm}
\label{table:domain_ordering_per_domain}

\end{table}

%% file: tables/kimi_vl.tex
\setlength{\columnsep}{8pt}

\begin{wraptable}{r}{0.5\textwidth}
\vspace{-4mm}
\small

\resizebox{\linewidth}{!}{%
\begin{tabular}{@{}l|ccc|c@{}}
\toprule
Method
& Math
& Medical
& Remote Sensing
& Mean \\
\midrule

Random (full)
  & 40.2
  & 46.5
  & 48.4
  & 45.0 \\

FFT
  & 40.0
  & 50.6
  & \textbf{51.5}
  & 47.4 \\

\expertlens
  & \textbf{41.5}
  & \textbf{51.4}
  & 51.0
  & \textbf{48.0} \\

\bottomrule
\end{tabular}%
}
\caption{\expertlens is competitive with full fine-tuning and outperforms random expert selection on \kimi.}
\label{table:kimi_vl_final_per_domain}
\vspace{-3mm}
\end{wraptable}

%% file: sections/conclusion.tex
\section{Conclusion}
In this work, we show that multimodal MoEs develop emergent semantic specialization across modalities, domains, and visual concepts, which can be exploited for efficient and effective adaptation. In this view, \expertlens is a structure-aware adaptation method: instead of introducing new parameters to adapt, it leverages the model's existing organization by identifying and updating experts already specialized for the target domain. To encourage further progress and model development, \textbf{we release all data, code, and trained models.}

%% file: sections/appendix.tex
\clearpage
\setcounter{page}{1}
\enablemaintoc
\appendix

\section{Appendix}
\label{appx}

\tableofcontents

\subsection{Preliminaries: Mixture-of-Experts Models}
\label{subsec:moes}
Mixture-of-Experts (MoE) models replace the dense feedforward network in a transformer block with a set of expert MLPs~\citep{lepikhin2020gshard, fedus2022switch}. Let $N$ denote the number of routed experts in each MoE layer, and let $\{E_i^{(\ell)}\}_{i=1}^{N}$ denote the experts in layer $\ell$. Given the hidden representation $h_t^{(\ell)}$ of token $t$, the learned router projection $W_r^{(\ell)}$ computes expert probabilities
\begin{equation}
    p_t^{(\ell)}
    =
    \operatorname{softmax}
    \left(
        W_r^{(\ell)}h_t^{(\ell)}
    \right),
\end{equation}
and selects the $K$ experts with the largest probabilities,
\begin{equation}
    \mathcal{T}_K^{(\ell)}(t)
    =
    \operatorname{TopK}
    \left(
        p_t^{(\ell)}
    \right).
\end{equation}
The layer output is the weighted combination of the selected experts,
\begin{equation}
    \operatorname{MoE}^{(\ell)}(h_t^{(\ell)})
    =
    \sum_{i\in\mathcal{T}_K^{(\ell)}(t)}
    \alpha_{t,i}^{(\ell)}
    E_i^{(\ell)}(h_t^{(\ell)}),
    \qquad
    \alpha_{t,i}^{(\ell)}
    =
    \frac{p_{t,i}^{(\ell)}}
    {\sum_{j\in\mathcal{T}_K^{(\ell)}(t)}
    p_{t,j}^{(\ell)}}.
\end{equation}
Some architectures additionally contain shared experts that are evaluated for every token. Sparse routing increases model capacity while improving throughput by evaluating only a small subset of the expert parameters for each token. In multimodal MoE models, both visual and textual tokens are processed by MoE layers, and routing is performed independently for each token. The resulting routing decisions may therefore depend on token modality, semantic content, and input domain. We say that expert $i$ in layer $\ell$ is \emph{activated} for token $t$ when $i\in\mathcal{T}_K^{(\ell)}(t)$. We distinguish its activation indicator
\begin{equation}
    a_{t,i}^{(\ell)}
    =
    \mathbbm{1}
    [i\in\mathcal{T}_K^{(\ell)}(t)]
    \label{eq:activation_indicator}
\end{equation}
from its routing mass
\begin{equation}
    m_{t,i}^{(\ell)}
    =
    \alpha_{t,i}^{(\ell)}
    a_{t,i}^{(\ell)}.
\end{equation}
Aggregating these quantities over tokens from a modality or domain provides a measure of the corresponding expert affinity. In this work, we study whether semantic specialization emerges in multimodal MoE models and whether this structure can be exploited for efficient, targeted adaptation.

\subsection{Additional Details on Visualizations}
\label{appx:visualization_details}
In this section, we provide additional details on the construction of the visualizations in \S\ref{sec:emergent_structure}. In \S\ref{appx:olo} we discuss Optimal Leaf Ordering and the datasets used in Figure~\ref{fig:image_text_ratio}. In ~\S\ref{appendix:knn} we provide additional details on the KNN purity analysis conducted in Figure~\ref{fig:knn_purity}. Finally, in \S\ref{appendix:tinyimagenet} we provide details on the semantic specialization analysis in Figure~\ref{fig:semantic_tinyimagenet}.

\subsubsection{Modality Specialization Addendum}
\label{appx:olo}

\textbf{Experimental Details.}
Image patches for this experiment were sourced from \breasthist~\citep{janowczyk2016deep}, \flowers~\citep{oxfordflowers}, \medmnist~\citep{yang2023medmnist}, \mathvista~\citep{mathvista}, \scienceqa~\citep{lu2022learn}, \cosyn~\citep{yang2025scaling}, \pathvqa~\citep{he2020pathvqa30000questionsmedical}, and text tokens from \textsc{MedQA} \citep{jin2020disease}, \textsc{PubMedQA} \citep{jin2019pubmedqadatasetbiomedicalresearch}, \textsc{GSM8K} \citep{gsm8k}, \textsc{ARC-Challenge} \citep{clark2018thinksolvedquestionanswering}, \openbookqa~\citep{mihaylov2018suitarmorconductelectricity}, \mmlu~\citep{mmlu}, and \legalbench~\citep{guha2023legalbench}.

\textbf{Optimal Leaf Ordering.}
In Fig.~\ref{fig:image_text_ratio}A, we visualize each expert's preference for image versus text tokens. To make structure among experts easier to see, we order experts within each layer according to the similarity of their router weights using \emph{optimal leaf ordering} (OLO). Importantly, this ordering is determined solely by the router weights and does not use the modality-specific activation statistics shown in the figure.

Recall from \S\ref{subsec:moes} that $W_r^{(\ell)}$ denotes the router projection for layer $\ell$. Let $w_i^{(\ell)}$ denote the row of $W_r^{(\ell)}$ corresponding to expert $i$. For each layer, we compute the pairwise cosine distance between expert router weights,
\[
D_{ij}^{(\ell)}
=
d_{\cos}\!\left(w_i^{(\ell)}, w_j^{(\ell)}\right).
\]
We use these distances to construct a hierarchical clustering dendrogram, i.e., a tree representing the nested grouping of experts by router-weight similarity.

A dendrogram does not uniquely determine the left-to-right ordering of its leaves, since the two children of any internal node can be exchanged without changing the underlying clustering. OLO resolves this ambiguity by selecting, among the orderings consistent with the dendrogram, a permutation $\pi_\ell$ that minimizes the total distance between adjacent experts:
\[
\pi_\ell
=
\arg\min_{\pi}
\sum_{k=0}^{N-2}
d_{\cos}\!\left(
w_{\pi(k)}^{(\ell)},
w_{\pi(k+1)}^{(\ell)}
\right).
\]
Thus, experts with similar router weights tend to appear near one another along the x-axis while preserving the hierarchical clustering structure.

After obtaining $\pi_\ell$, the expert shown at x-axis position $k$ is $i=\pi_\ell(k)$. We then compute its image-fraction score using the activation indicator $a_{t,i}^{(\ell)}$ defined in Eq.~\ref{eq:activation_indicator}. Let $\mathcal{I}_d$ and $\mathcal{X}_d$ denote the sets of image and text tokens, respectively, sampled from domain $d$. We define
\[
f_{\mathrm{img}}^{(\ell)}(i)
=
\frac{
\sum_d \sum_{t\in\mathcal{I}_d} a_{t,i}^{(\ell)}
}{
\sum_d
\left(
\sum_{t\in\mathcal{I}_d} a_{t,i}^{(\ell)}
+
\sum_{t\in\mathcal{X}_d} a_{t,i}^{(\ell)}
\right)
}.
\]
Because image and text tokens are sampled equally, a score near $0.5$ indicates no strong modality preference, while values approaching $1$ or $0$ indicate increasingly image- or text-specific activation, respectively.

OLO therefore affects only the horizontal ordering of experts in Fig.~\ref{fig:image_text_ratio}A; each expert's modality preference is computed independently from its observed top-$K$ routing behavior.

\subsubsection{kNN Purity}
\label{appendix:knn}

We use k-nearest-neighbor (kNN) purity to quantify which factors best explain the local structure of token representations. Given the same activation vectors, we assign alternative categorical labels to each sample---e.g., semantic dataset or part of speech for text tokens, and semantic dataset, color, or spatial position for image patches---and measure how strongly nearby representations share each label. Higher kNN purity indicates that the corresponding factor better explains the local organization of the activation space.

\mypar{Data.}
\label{appx:knndata}
For image patches, we sample from 12 VQA datasets spanning a diverse set of visual domains: \breasthist~\citep{janowczyk2016deep}, \flowers~\citep{oxfordflowers}, \medmnist~\citep{yang2023medmnist}, \mathvista~\citep{mathvista}, \scienceqa~\citep{lu2022learn}, \cosyn~\citep{yang2025scaling}, \aiid~\citep{ai2d}, \mmk~\citep{mmk}, \geomverse~\citep{kazemi2023geomverse}, \pathvqa~\citep{he2020pathvqa30000questionsmedical}, \vqarad~\citep{vqarad}, and \charxiv~\citep{charxiv}. Each dataset defines one category when evaluating organization by semantic dataset.

For text tokens, we sample from 12 text-based question-answering datasets spanning diverse domains and linguistic content: \medqa~\citep{jin2020disease}, \pubmedqa~\citep{jin2019pubmedqadatasetbiomedicalresearch}, \medmcqa~\citep{pmlr-v174-pal22a}, \gsm~\citep{gsm8k}, \metamath~\citep{yu2024metamath}, \arcchallenge~\citep{clark2018thinksolvedquestionanswering}, \openbookqa~\citep{mihaylov2018suitarmorconductelectricity}, \mmlu~\citep{mmlu}, \triviaqa~\citep{joshi2017triviaqa}, \bbh~\citep{srivastava2023beyond}, \finqa~\citep{chen2021finqa}, and \legalbench~\citep{guha2023legalbench}. For the part-of-speech analysis, tokens are assigned to one of 12 categories:
\[
\{\mathrm{VERB}, \mathrm{NOUN}, \mathrm{PRON}, \mathrm{ADJ},
\mathrm{ADV}, \mathrm{ADP}, \mathrm{CONJ}, \mathrm{DET},
\mathrm{NUM}, \mathrm{PRT}, ``.'', ``?''\}.
\]
We balance the number of samples across categories for each comparison so that purity is not driven by differences in category frequency.

\mypar{kNN Purity.}
Let $\mathbf{x}_i \in \mathbb{R}^D$ denote the activation vector of token or image patch $i$, with categorical label $y_i$. For each sample, we find its $k$ nearest neighbors in the original $D$-dimensional activation space using Euclidean distance,
\[
d(\mathbf{x}_i,\mathbf{x}_j)
=
\lVert \mathbf{x}_i-\mathbf{x}_j\rVert_2,
\]
excluding the sample itself. Let $\mathcal{N}_k(i)$ denote the indices of these $k$ neighbors. We define the local purity of sample $i$ as
\[
p_k(i)
=
\frac{1}{k}
\sum_{j\in\mathcal{N}_k(i)}
\mathbb{I}[y_j=y_i],
\]
i.e., the fraction of its nearest neighbors that share its categorical label.

For a labeling scheme containing $N$ sampled representations, the reported kNN purity is the mean local purity across samples:
\begin{equation*}
P_k
=
\frac{1}{N}
\sum_{i=1}^{N} p_k(i)
=
\frac{1}{Nk}
\sum_{i=1}^{N}
\sum_{j\in\mathcal{N}_k(i)}
\mathbb{I}[y_j=y_i],
\end{equation*}
where $\mathbb{I}[\cdot]$ is the indicator function. We report this quantity for multiple values of $k$ in Fig.~\ref{fig:knn_purity}.

\subsubsection{Semantic Specialization Analysis}
\label{appendix:tinyimagenet}

We quantify whether semantically related object classes recruit similar sets of experts using the \tin experiments described in \S\ref{subsec:semantic_specialization}. For each MoE layer and patch-count threshold $m$, we first construct a set of experts associated with each object class. An expert is considered active for an image if it is routed at least $m$ of that image's patch tokens, and is included in the class-level expert set $S_c$ if it is active for at least $60\%$ of images from class $c$. We evaluate multiple thresholds $m\in\{3,5,8,12,16\}$ to ensure that the analysis is not sensitive to a particular definition of expert activation.

\mypar{Expert-set overlap.}
For two classes $a$ and $b$, we measure the overlap between their associated expert sets using the Jaccard similarity,
\[
J(a,b)
=
\frac{|S_a\cap S_b|}
{|S_a\cup S_b|}.
\]
A larger $J(a,b)$ therefore indicates that the two classes consistently recruit more of the same experts.

Raw Jaccard overlap can vary across classes because some classes may recruit expert sets that overlap broadly with many others. To account for this, we standardize each pairwise overlap relative to the typical overlap of the two classes with the remainder of the class set. For an analysis containing $N$ classes, we define
\[
\mu_{a,b}
=
\frac{1}{2(N-2)}
\left(
\sum_{k\neq a,b}J(a,k)
+
\sum_{k\neq a,b}J(b,k)
\right),
\]
which measures the average overlap of classes $a$ and $b$ with all other classes. We then compute
\[
Z(a,b)
=
\frac{J(a,b)-\mu_{a,b}}{\sigma},
\]
where $\sigma$ is the standard deviation of all off-diagonal entries in the pairwise Jaccard matrix.

Thus, $Z(a,b)>0$ indicates that classes $a$ and $b$ share more experts with each other than they typically share with other classes, while larger values indicate increasingly strong shared expert specialization.

\subsection{ExpertLens Classification}
\label{appx:el_classification}
We provide additional details for \expertlens, our data-free method for identifying domain-specialized experts in multimodal MoE models. For each MoE expert, we project its router weight through the language model head and extract the $10$ highest-scoring vocabulary tokens. For each domain, we prompt \qwen to classify each token as belonging to that domain or not using deterministic greedy decoding. We include all prompts used in \S\ref{appx:prompts}. 

\begin{figure}[h]
\centering
\includegraphics[width=0.9\linewidth]{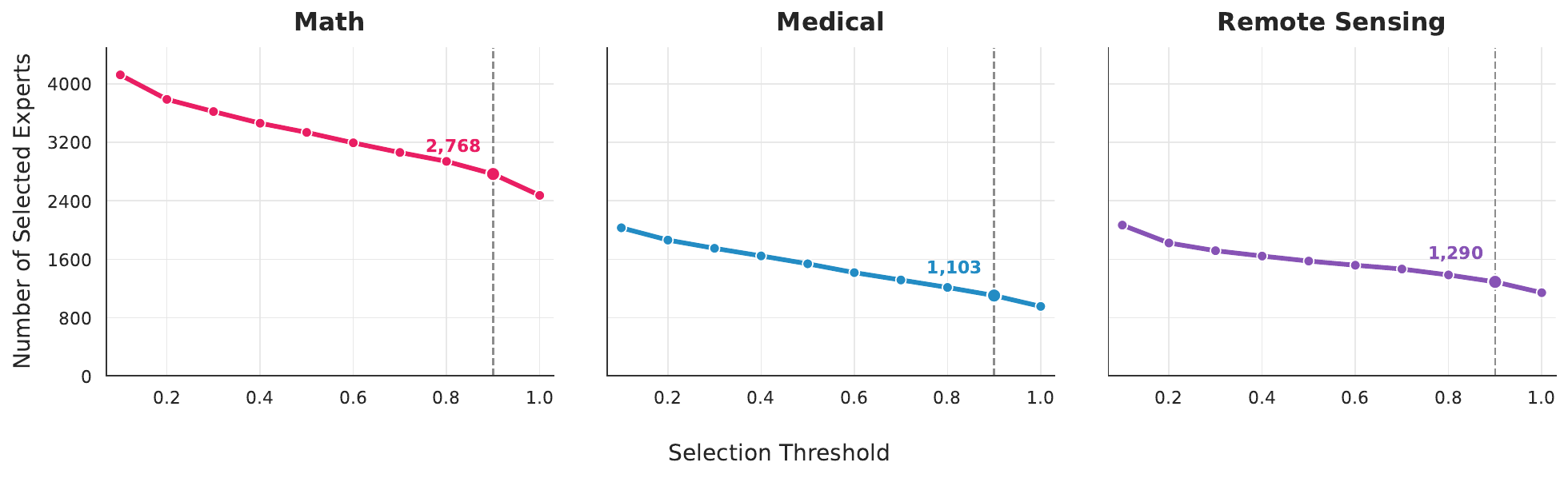}
\vspace{-3mm}
\caption{\textbf{Sensitivity to score threshold.} We vary the \expertlens selection threshold ($S_{\mathrm{EL}}$) and report the resulting number of selected experts for each domain in Qwen3-VL-30B-A3B. Expert counts decrease smoothly as the threshold becomes more stringent. Our default threshold of \(S_{\mathrm{EL}} \geq 0.9\) (dashed line) selects 2,768 math, 1,103 medical, and 1,290 remote-sensing experts.}
\vspace{-1mm}
\label{fig:sensitivity}
\end{figure}

\mypar{Expert selection is stable around the chosen score threshold.} We examine the sensitivity of \expertlens expert cohorts to the selection threshold \(S_{\mathrm{EL}}\) in Figure~\ref{fig:sensitivity}. As expected, increasing the threshold progressively reduces the number of experts selected for each domain. Importantly, cohort sizes vary smoothly across thresholds, with no sharp transition around our default value of \(S_{\mathrm{EL}} \geq 0.9\). At this threshold, ExpertLens selects 2,768, 1,103, and 1,290 experts for math, medical, and remote sensing, respectively. We use this same threshold across domains rather than tuning it independently for each task.

\input{tables/noisy_classifications}

\mypar{Limitation: noisy classifications.} We now discuss a primary limitation of \expertlens: susceptibility to noise in token-level semantic classification. Since \expertlens determines expert specialization from the semantic relevance of decoded vocabulary tokens, its predictions inherit errors from this token-level classification step. As shown in Table~\ref{tab:expertlens_noisy_tokens}, expert readouts can include ambiguous or non-semantic tokens, such as token fragments, punctuation, multilingual fragments, and control tokens. The \qwen model used for classification occasionally incorrectly labels these tokens as pertaining to a specific domain. Such errors introduce noise into the inferred expert labels and represent a primary failure mode of our approach.

\begin{figure}[h]
    \centering
    \includegraphics[width=0.99\linewidth]{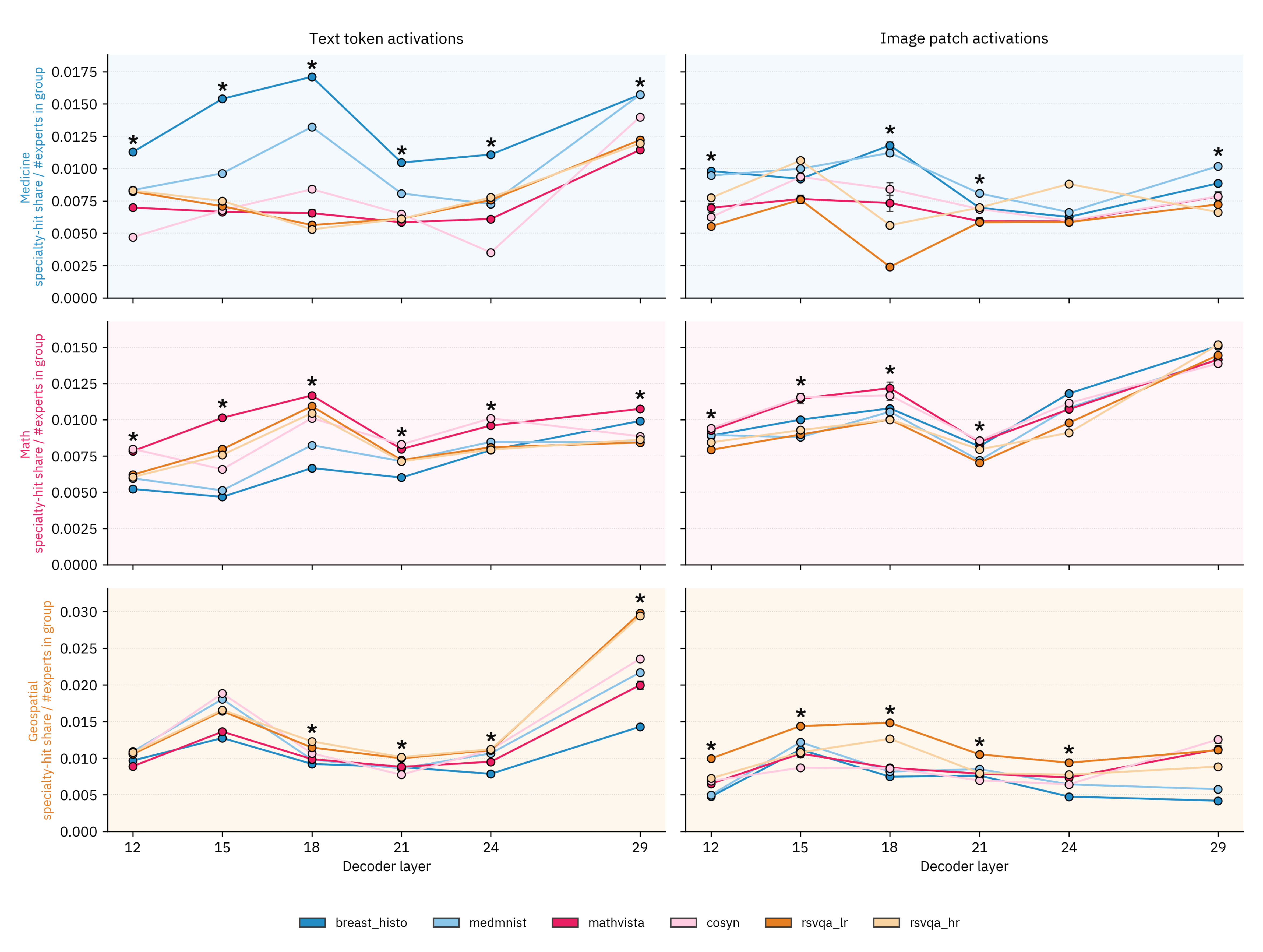}
    \caption{\textbf{ExpertLens-derived expert specialization persists across layers.}
We extend the analysis from Fig.~\ref{fig:expertlens_viewing_ab}A across decoder layers of \qwen. Rows correspond to \expertlens-identified medical, math, and remote-sensing expert cohorts, and columns show routing for text tokens and image patches. Curves report the normalized dataset-hit ratio per expert for each of the six evaluation datasets; asterisks mark layers where at least one of the top-two preferred datasets belongs to the expert cohort's corresponding domain. Across layers, expert cohorts preferentially activate on inputs from their matching domain, showing that the domain alignment observed in Fig.~\ref{fig:expertlens_viewing_ab}A is not specific to a single layer.}
    \label{fig:expertlens_across_layers}
\end{figure}

\mypar{ExpertLens specialization persists across layers.}
Figure~\ref{fig:expertlens_across_layers} extends the routing analysis from Fig.~\ref{fig:expertlens_viewing_ab}A, which visualizes layer 18, to multiple decoder layers. For each layer, we independently identify medical, math, and remote-sensing experts using \expertlens and measure their routing preferences over the same six datasets. The domain correspondence observed at layer 18 persists broadly throughout the decoder: medical experts preferentially activate on medical datasets, math experts on mathematical datasets, and remote-sensing experts on remote-sensing datasets. Although the strength of this preference varies across layers and modalities, the overall pattern is consistent, indicating that \expertlens recovers domain-specialized experts throughout the model rather than only at a particular layer.

\clearpage
\subsection{Additional Analyses}
We now provide additional analyses to supplement those in the main paper. We evaluate the distribution of routing scores in \S\ref{appx:sharpness}, provide additional visualizations for modality preferences (\S\ref{appx:modality}), domain specificity (\S\ref{appx:purity}, \S\ref{appx:tsne}), and semantic specialization (\S\ref{appx:addtl_semantic_vis}). Finally, we explore the stability of routing preferences with \expertlens post-training in \S\ref{appx:routing_stability}.

\begin{figure}
    \centering
\includegraphics[width=\linewidth]{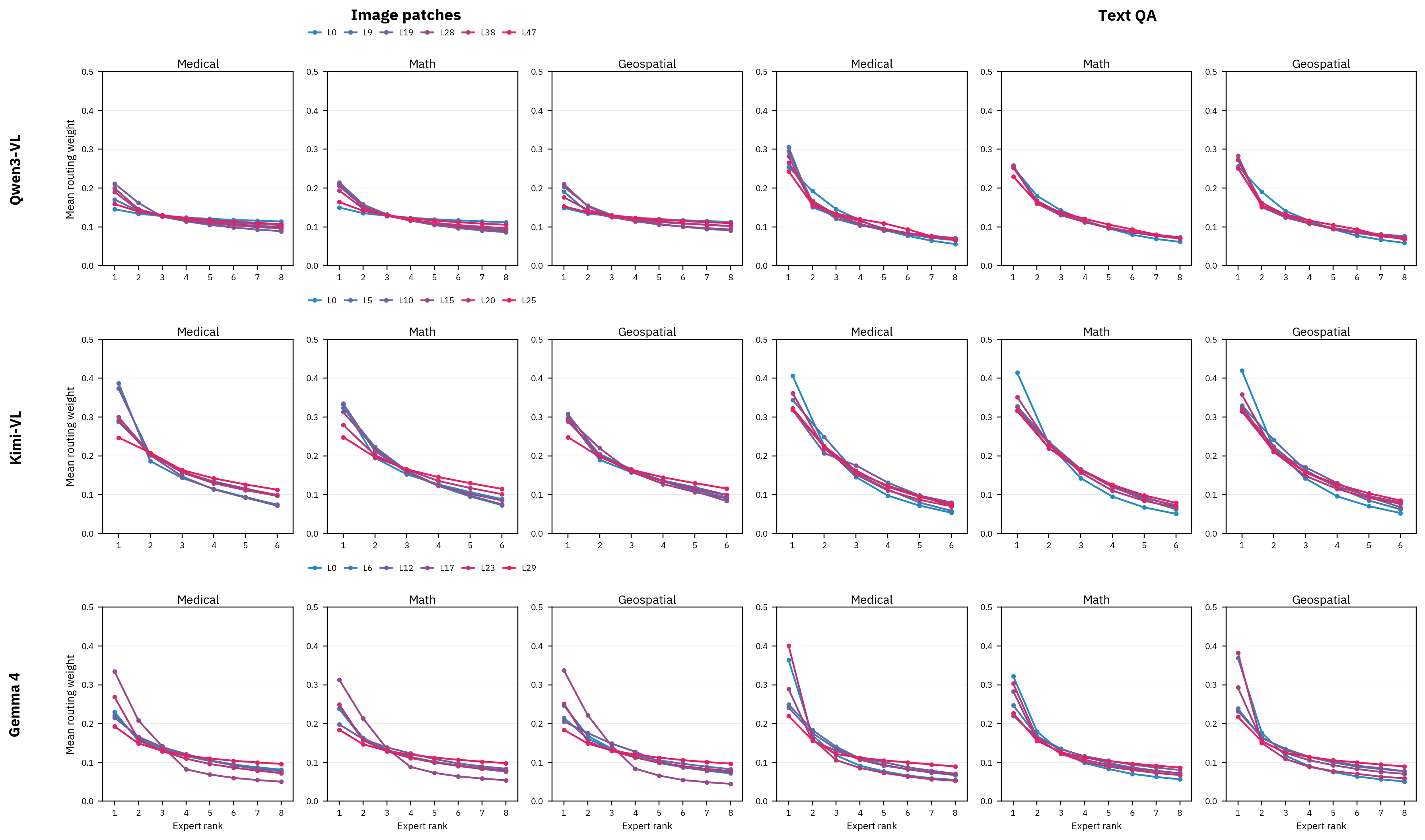}
    \caption{Mean routing weights for the top-k selected experts from each model, on image patches and text tokens. The difference between the top-1 and top-k expert weight is larger for text tokens than image patches in all models. Routing distributions across domains is similar within each modality. Overall, no model exhibits a mean top-1 routing weight over 0.5 or a top-k routing weight below 0.05.}
    \label{fig:appendix peaky}
\end{figure}

\subsubsection{Routing Sharpness Analysis}
\label{appx:sharpness}

We additionally examine how routing mass is distributed among the top-$K$ selected experts. Figure~\ref{fig:appendix peaky} plots the mean routing weight assigned to experts at each rank, separately for image patches and text tokens across medical, math, and remote sensing domains. Across all three models, routing weights decrease smoothly with expert rank rather than collapsing onto a single expert, indicating that routing remains distributed across several selected experts. Text tokens generally exhibit sharper routing than image patches, with a larger gap between the top-ranked expert and the remaining experts. In contrast, routing profiles are broadly similar across domains within each modality, suggesting that routing sharpness is driven more strongly by modality than by domain. Overall, no model assigns more than $0.5$ mean routing weight to its top-ranked expert, and even the lowest-ranked selected experts retain non-negligible routing mass, indicating that the observed expert specialization does not arise from near-deterministic routing to a single expert.

\subsubsection{Additional Modality Specificity Analysis}
\label{appx:modality}
\begin{figure}[h]
    \centering
\includegraphics[width=0.65\linewidth]{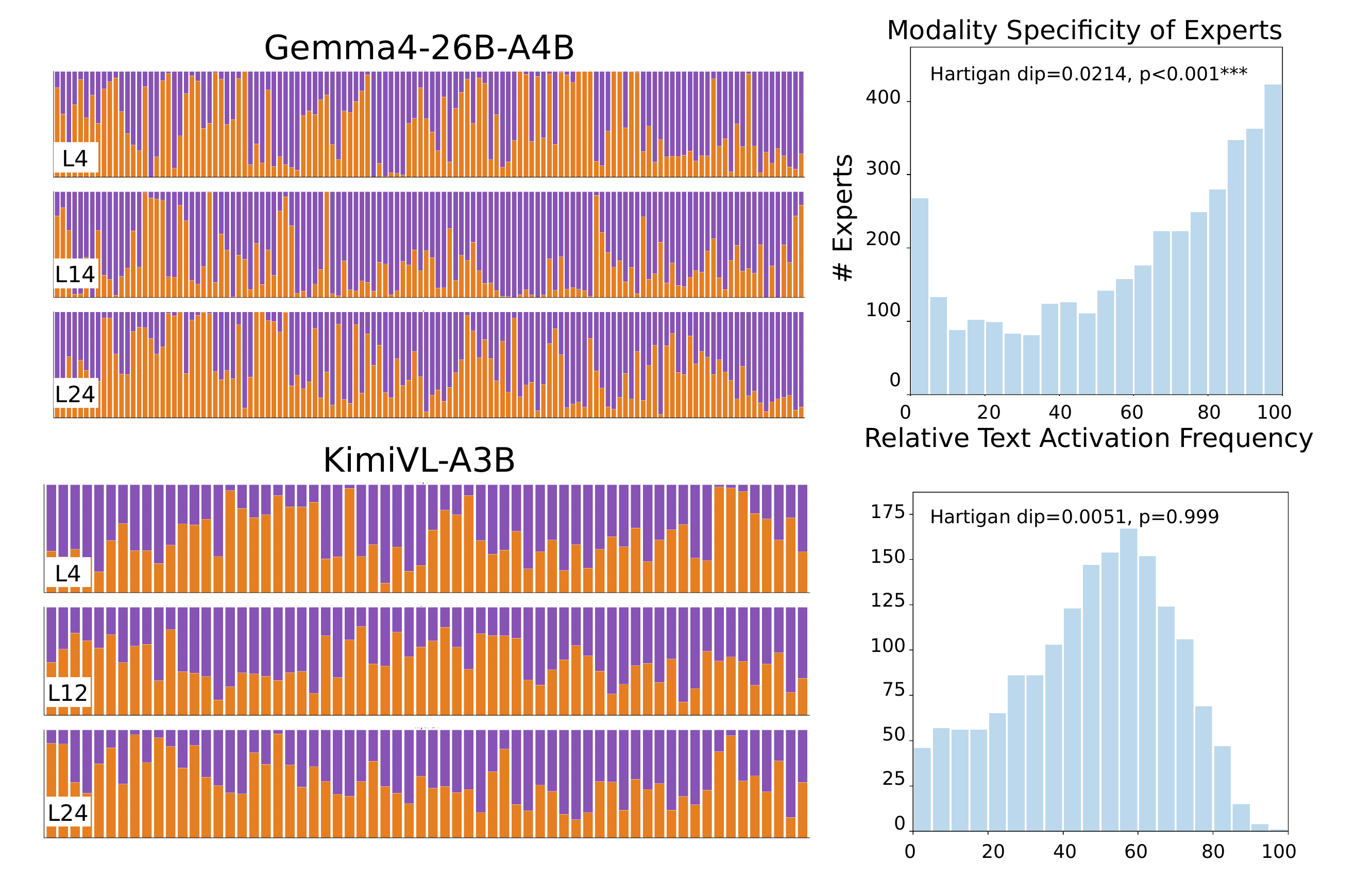}
    \caption{Expanding on Fig.\ref{fig:image_text_ratio}, \gemma{} shows a strong modality-preference with a bimodal frequency histogram, while \kimi{} with fewer net experts appears to consist of mostly neutral experts. }
    \label{fig:modality_kimi_gemma}
\end{figure}

We supplement our analysis from \S\ref{sec:expertspecialization_modality} with similar plots for \kimi and \gemma in Figure~\ref{fig:modality_kimi_gemma}. On the left, we show the fraction of top-K activations arising from image patches versus text tokens, organized by OLO (\S\ref{appx:olo}). On the right, we show the number of experts per distinct frequency bins of text-to-image hits ratio. \gemma{} displays a clear bimodal distribution of expert specialization, with a Hartigan dip value of 0.0214 ($p<0.001$), whereas \kimi{} appears to consist of experts without strong modality preference. Here, note that \gemma{} and \qwen{} have 128 experts per layer, while \kimi{} has only 64, possibly explaining its lack of modality-specific experts.
These results build on our findings from
\S\ref{sec:expertspecialization_modality} and suggest modality preferences are not specific to \qwen.

\begin{figure}[h]
    \centering
\includegraphics[width=\linewidth]{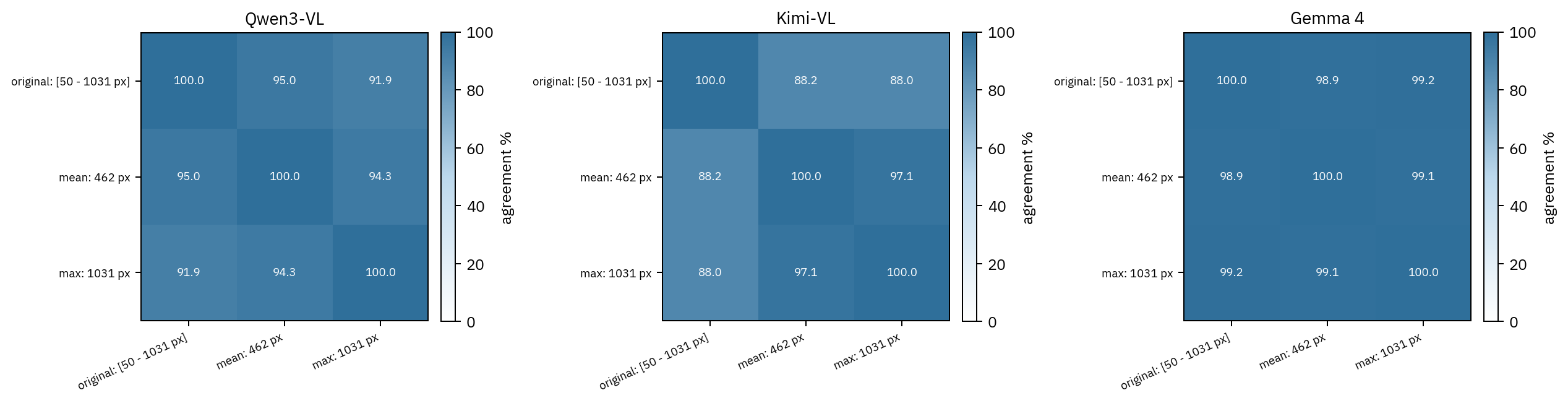}
    \caption{Robustness to image scale. We repeat the experiment in \S\ref{sec:expertspecialization_modality} but vary the scale of images.
Specifically, we fix the total number of patches per domain and vary only image width
(\texttt{original}: $[50,1031]$\,px;
\texttt{mean}: $462$\,px;
\texttt{max}: $1031$\,px),
so smaller images contribute fewer patches each and we sample from more images. We then count the number of image-leaning versus text-leaning experts in the different experimental runs.
Cells show the fraction of experts that retain the same $\ge 50\%$ image/text side across a pair of settings.
High agreement across scale changes implies modality specialization is stable under image scale change.}
    \label{fig:appendix scale}
\end{figure}

\mypar{Modality preferences are robust to image scale.}
We test whether the modality specialization observed in \S~\ref{sec:expertspecialization_modality} is sensitive to image scale. For each model, we repeat the analysis while varying image width across the original resolution distribution, the dataset mean resolution, and the maximum resolution, while keeping the total number of sampled image patches per domain fixed. We then measure, for each pair of settings, the fraction of experts that remain on the same side of the $50\%$ image/text activation boundary. Figure~\ref{fig:appendix scale} shows high agreement across scale settings for all three models, indicating that expert modality preference is largely unchanged by image resolution. Thus, the modality specialization we observe is unlikely to be an artifact of image scale or the number of patches contributed by individual images.

\subsubsection{Additional kNN Purity Analysis}
\label{appx:purity}
\begin{figure}
    \centering
\includegraphics[width=0.78\linewidth]{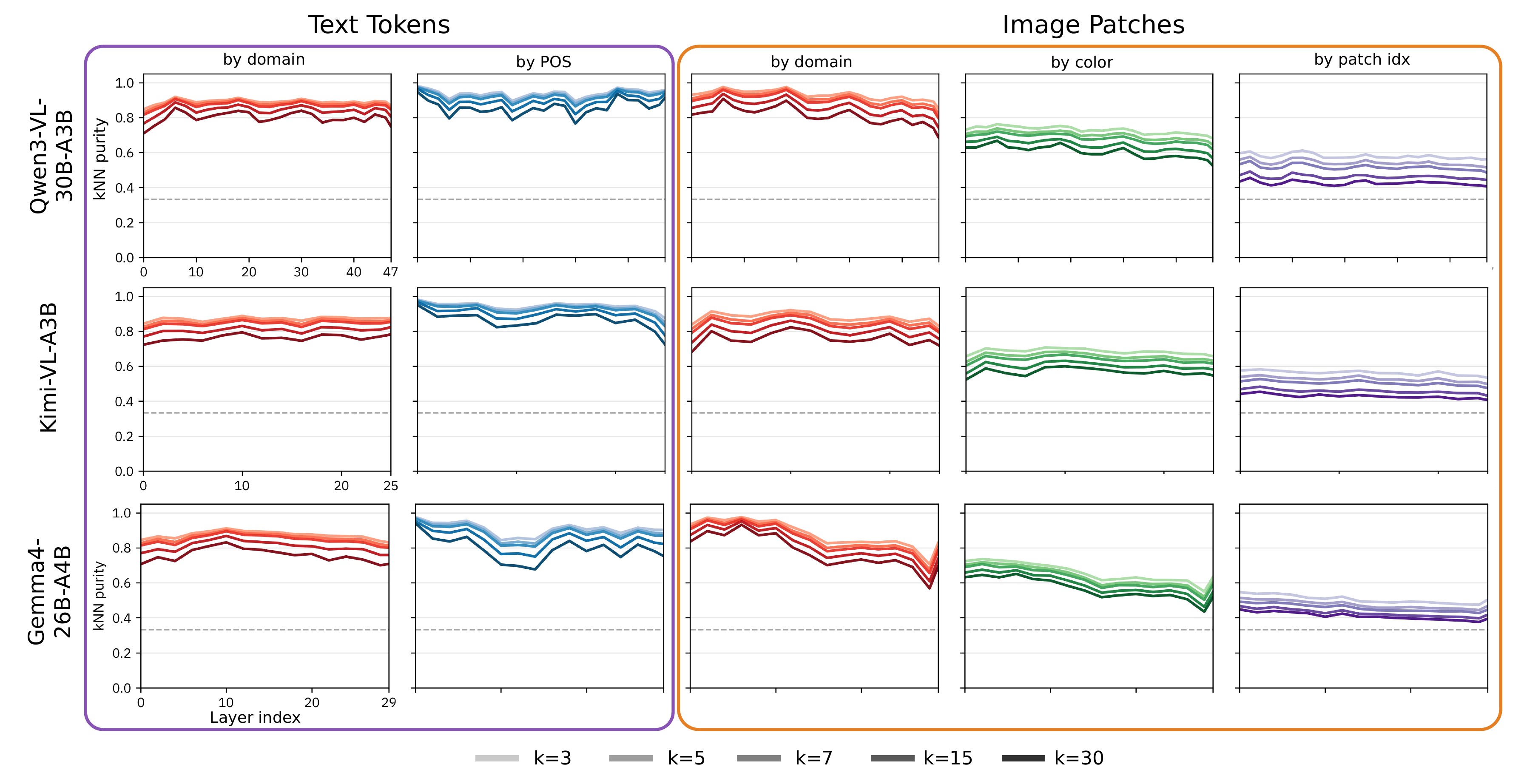}
    \caption{Expanding on Fig.\ref{fig:knn_purity}, which showed that grouping by dataset showed highest kNN purity for image activations, while grouping by POS showed highest kNN purity for text tokens. Clustering by 3 domains (medical, math, remote sensing) with images from 2 datasets per domain, we see that grouping by domain is still the most kNN-pure method in image activations, while POS narrowly exceeds grouping by domain in text tokens. All kNN purity values are higher due to the fewer number of classes. The gray dotted line indicates chance kNN purity.}
    \label{fig:knn purity by domain appendix}
\end{figure}

We now expand on our domain-specificity analysis in ~\S\ref{subsec:domain_specificity}. Specifically, Figure~\ref{fig:knn_purity} showed that grouping by dataset led to the highest purity for image activations, while part-of-speech showed the highest purity for text tokens. In Figure~\ref{fig:knn purity by domain appendix} we repeat this analysis with clustering by domain. Specifically, we consider our three domains (math, medical, remote sensing) and sample from 2 datasets per domain. Grouping by domain is still the most kNN-pure method in image activations, while POS narrowly exceeds grouping by domain in text tokens. All kNN purity values are higher due to the fewer number of classes. Additionally, we find that both groupings are higher than chance-grouping (shown by the gray dotted line).

\subsubsection{Additional Semantic Specialization Visualizations.}
\label{appx:addtl_semantic_vis}
\begin{figure}[h]
    \centering
\includegraphics[width=0.6\linewidth]{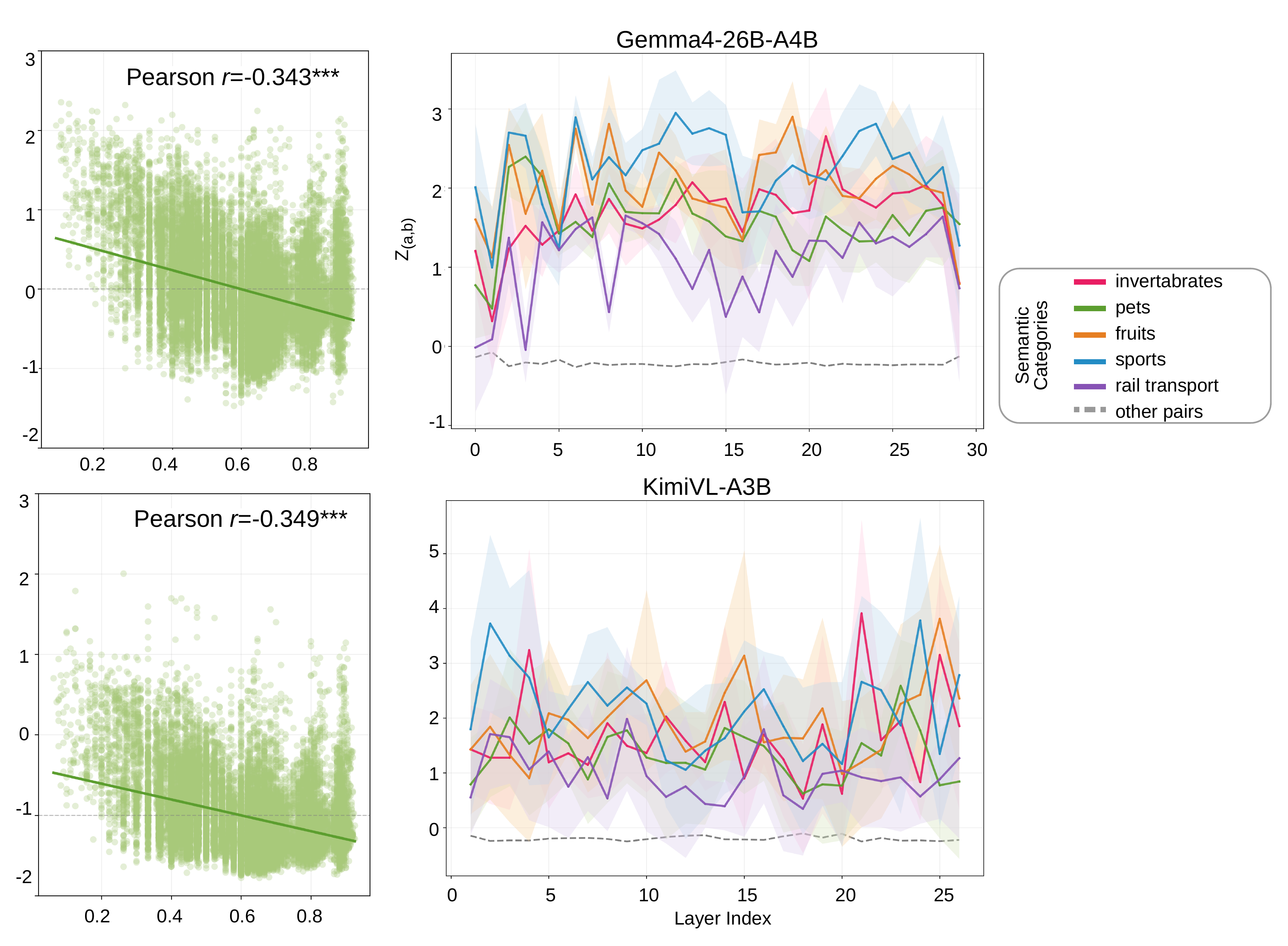}
    \caption{Expanding on Fig. \ref{fig:semantic_tinyimagenet}, object level categorization can be observed in \kimi{} and \gemma. The left hand for each plot shows Z(a, b) plotted against
the semantic distance between object class pairs (x axis). Semantically unrelated object classes share fewer
experts, as with \qwen. The right hand side for each plot shows the category level expert specialization compared to unrelated category pairs. Images from objects in the same semantic category share more experts than any two unrelated object classes.}
    \label{fig:appendix tinyimagenet}
\end{figure}

We extend our analysis in ~\S\ref{subsec:semantic_specialization} to \kimi and \gemma. Specifically, in Figure~\ref{fig:appendix tinyimagenet} we find that both \gemma and \kimi illustrate the same behavior as \qwen (with Pearson coefficients of $-0.343$ and $-0.349$ respectively, comparable to \qwen's $-0.380$ (left). Moreover, the right column in Figure~\ref{fig:appendix tinyimagenet} shows that semantically related classes elicit similar experts, with all pairs of related classes above the gray baseline.

\subsubsection{Routing Stability Analysis}
\label{appx:routing_stability}
Although ExpertLens updates the router throughout continual adaptation (Table~\ref{table:freezing_ablations}), its routing structure remains highly stable. Consecutive checkpoints preserve $92.0-94.0\%$ top-8 expert-set overlap, measured as the Jaccard similarity between the highest-probability experts selected for each input and layer. After completing all three training stages, the final model retains $90.1\%$ overlap with the pretrained router, with similarly high retention across Math ($90.2\%$), Medical ($90.1\%$), and Remote Sensing ($89.9\%$) inputs. Thus, training the router enables adaptation without substantially disrupting the model's pretrained routing structure or causing routing collapse.

\subsubsection{Additional t-SNE Projections}
\label{appx:tsne}
\begin{figure}[h]
    \centering
    \includegraphics[width=0.65\linewidth]{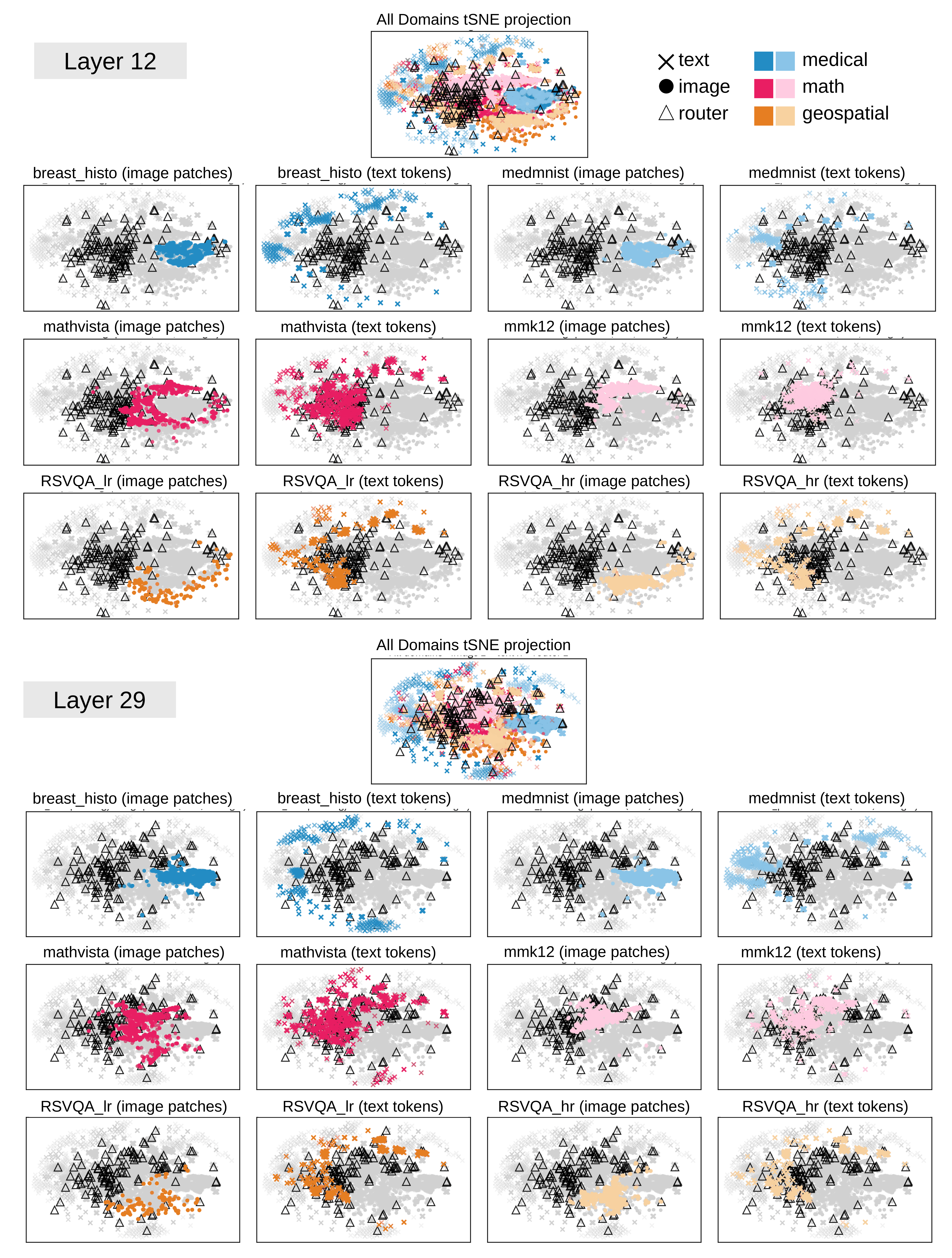}
    \caption{\textbf{t-SNE projections for additional layers of \qwen}. We visualize layers 12 and 29 of \qwen, supplementing the plot included in the main paper. Across various layers we find that visual activations exhibit stronger domain structure than textual activations.}
    \label{fig:tsne_qwen_other_layers}
\end{figure}

\begin{figure}[t]
    \centering
    \includegraphics[width=0.65\linewidth]{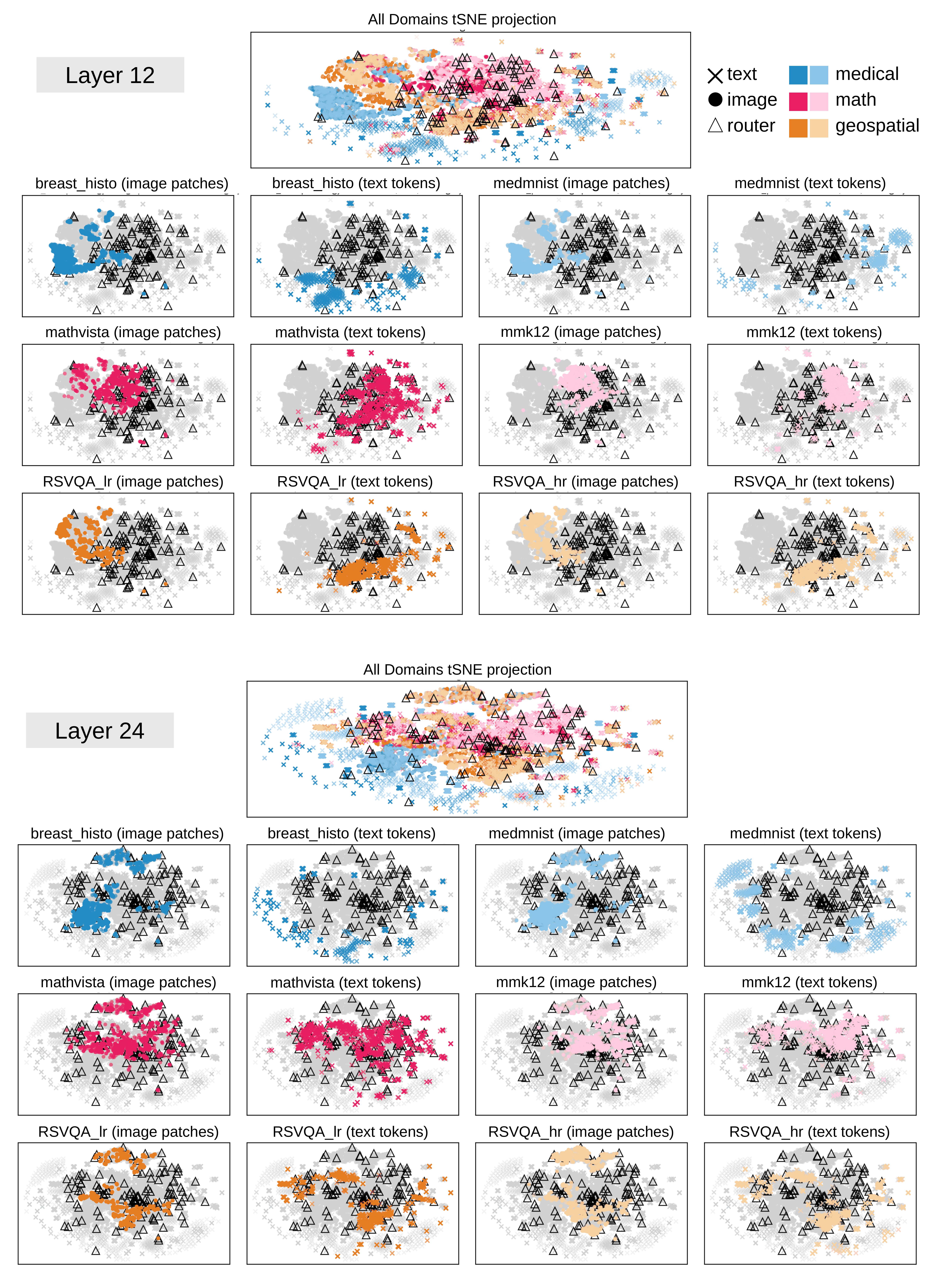}
    \caption{\textbf{t-SNE projections for \gemma}. We visualize Layers 12 and 24 and find image patches are more tightly clustered than text tokens.}
    \label{fig:tsne_gemma}
\end{figure}

\begin{figure}[t]
    \centering
    \includegraphics[width=0.65\linewidth]{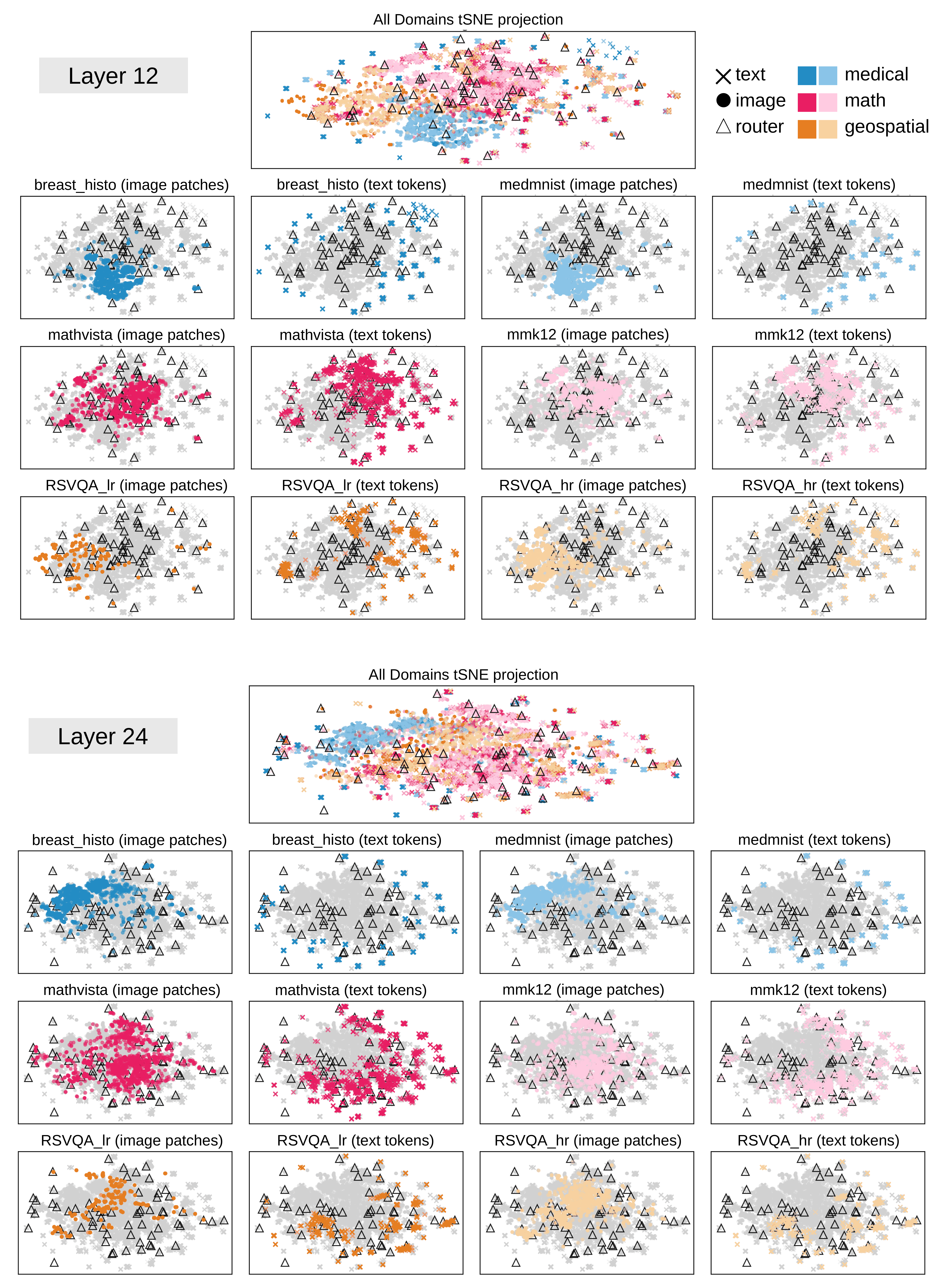}
    \caption{\textbf{t-SNE projections for \kimi}. We visualize Layers 12 and 24 and find image patches are more tightly clustered than text tokens.}
    \label{fig:tsne_kimi}
\end{figure}

To supplement our analysis in \S\ref{subsec:domain_specificity}, we provide additional t-SNE projections. Specifically, we visualize additional layers for \qwen in Figure~\ref{fig:tsne_qwen_other_layers}, and provide visualizations for \kimi in Figure~\ref{fig:tsne_kimi} and \gemma in Figure~\ref{fig:tsne_gemma}. Across all visualizations, we find that projections of image patch activations are highly clustered by domain, while textual activations are comparatively more diffuse.

\clearpage

\subsection{Experimental Details}
\label{appx:experiment_details}

\subsubsection{Training Details}
\label{appx:training_details}

\mypar{Training Data and Rewards.} We train on $4{,}000$ examples per domain, sampled from the corresponding training splits. For math, all samples are from \mmk~\citep{mmk}. For medical, we use all $1{,}793$ samples from \vqarad~\citep{vqarad} and $2{,}207$ examples from \pathvqa~\citep{he2020pathvqa30000questionsmedical}. For remote sensing, we sample $2{,}000$ examples each from \rsvqalr~\citep{rsvqa} and \rsvqahr~\citep{rsvqa}. We adopt the system prompt and reward functions from VERO~\citep{vero}.

\mypar{Implementation.} We use \qwen~as our base multimodal MoE and GSPO~\citep{gspo} as our post-training algorithm. All experiments are conducted on $4$ H200 GPUs for $200$ steps per domain, with a batch size of $32$, group size of $5$, and learning rate of $1\times10^{-5}$. We use verl~\citep{verl} for training. Additional hyperparameters are provided in Table~\ref{tab:qwen3_hyperparams}. We show results for three seeds, with shaded regions indicating $95\%$ confidence intervals.

\subsubsection{Evaluation Details}

We evaluate all models using lmms-eval~\citep{lmms_eval2024} with the vLLM inference backend. We use zero-shot evaluation and sample responses with recommended sampling parameters for \qwen: temperature $0.7$, top-$p$ 
$0.8$, top-$k$ $20$, a presence penalty of $1.5$, and a repetition penalty of 
$1.0$, allowing up to 4,096 generated tokens. We use identical prompts, image 
preprocessing, decoding parameters, and evaluation seeds for all methods. 

\mypar{Evaluation Benchmarks.} We evaluate on benchmarks spanning all three domains. For medical, \pathvqa~\citep{he2020pathvqa30000questionsmedical}
pairs pathology images with open-ended and yes/no questions about clinical
findings. \vqarad~\citep{vqarad} contains clinician-authored open- and
closed-ended questions over radiographs, CT scans, and MRI images, covering
anatomy, imaging modality, and abnormalities. \slake~\citep{slake} spans multiple imaging modalities and body regions, with questions requiring both visual recognition and medical knowledge.

For mathematical reasoning, \mathvista~\citep{mathvista} covers diverse visual
mathematics tasks, including arithmetic, geometry, logical reasoning, and
interpretation of charts, tables, and scientific figures.
\mathvision~\citep{mathvision} contains challenging visual problems drawn from
real mathematics competitions, spanning 16 mathematical disciplines and five
difficulty levels. \mmk~\citep{mmk} comprises manually collected K--12
multiple-choice problems.

For remote sensing, \hrvqa~\citep{hrvqa} uses high-resolution aerial imagery
to test counting, scene recognition, spatial understanding, and fine-grained
object attributes such as size, color, and shape.
\rsvlmqa~\citep{rsvlmqa} combines remote-sensing imagery with questions about
captions, semantic content, spatial relationships, and object counting.
\rsvqalr~\citep{rsvqa} uses lower-resolution Sentinel-2 imagery to evaluate
object presence, counting, comparison, and rural-versus-urban classification,
whereas \rsvqahr~\citep{rsvqa} uses very-high-resolution aerial orthophotos to
test fine-grained object presence, counting, comparison, and area estimation.

\subsubsection{Domain Ordering Experimental Variability}
In Table~\ref{table:domain_ordering_stdev} we report standard deviations across three seeds for the domain ordering results shown in Table~\ref{table:domain_ordering_per_domain}. We find that \expertlens generally exhibits lower intra-seed variation, as well as lower variability per-ordering.

\begin{table}[h]
\centering
\small

\resizebox{0.9\linewidth}{!}{%
\begin{tabular}{
  @{}
  ll
  c@{\hspace{3em}}
  c@{\hspace{3em}}
  c@{\hspace{3em}}
  |
  c@{\hspace{2.8em}}
  @{}
}
\toprule
Domain Ordering
& Method
& Math
& Medical
& Remote Sensing
& Mean \\
\midrule

\multirow{2}{*}{
  Math {\scriptsize$\rightarrow$}
  Med.\ {\scriptsize$\rightarrow$}
  Rem.\ Sens.
}
& FFT
  & 1.3
  & \textbf{0.7}
  & 0.4
  & 0.8 \\
& \expertlens
  & \textbf{0.4}
  & \textbf{0.7}
  & \textbf{0.2}
  & \textbf{0.4} \\

\cmidrule(lr){1-6}

\multirow{2}{*}{
  Med.\ {\scriptsize$\rightarrow$}
  Math {\scriptsize$\rightarrow$}
  Rem.\ Sens.
}
& FFT
  & \textbf{0.7}
  & 0.9
  & \textbf{0.2}
  & \textbf{0.6} \\
& \expertlens
  & 2.0
  & \textbf{0.2}
  & 0.5
  & 0.9 \\

\cmidrule(lr){1-6}

\multirow{2}{*}{
  Rem.\ Sens.\ {\scriptsize$\rightarrow$}
  Med.\ {\scriptsize$\rightarrow$}
  Math
}
& FFT
  & 1.6
  & 1.0
  & \textbf{0.2}
  & 0.9 \\
& \expertlens
  & \textbf{0.1}
  & \textbf{0.1}
  & 0.3
  & \textbf{0.2} \\

\bottomrule
\end{tabular}%
}

\vspace{-2mm}
\caption{\textbf{Domain ordering experimental variability.} We report standard deviations across three seeds for our domain ordering experiments.}
\vspace{-5mm}
\label{table:domain_ordering_stdev}
\end{table}

\clearpage

\subsubsection{Hyperparameters}
\begin{table}[h]
\centering
\small
\begin{tabular}{@{} l r @{}}
\toprule
\textbf{Hyperparameter} & \textbf{Value} \\
\midrule
Training objective                  & GSPO \\
Training steps per domain           & 200 \\
Batch size                          & 32 \\
Group size                          & 5 \\
Learning rate                       & $1\times10^{-5}$ \\
Optimizer                           & AdamW \\
Weight decay                        & 0 \\
LR schedule                         & Constant \\
Maximum prompt length               & 1,024 \\
Maximum response length             & 2,000 \\
KL coefficient                      & 0 \\
Clipping range                      & $[0.20,\,0.28]$ \\
Temperature                         & 1.0 \\
Top-p                               & 1.0 \\
Top-k                               & Disabled \\
Repetition / presence penalty       & 1.0 / 1.5 \\
Maximum image pixels                & 401,408 \\
ExpertLens selection threshold      & $\geq 0.9$ \\
Training and rollout dtype          & bfloat16 \\
Rollout engine                      & vLLM \\
\bottomrule
\end{tabular}
\caption{Hyperparameters for the Qwen3-VL experiments.}
\label{tab:qwen3_hyperparams}
\end{table}

\subsection{Prompts}
\label{appx:prompts}

\input{prompts/EL_system}
\vspace{3mm}
\input{prompts/EL_classification}

%% file: tables/noisy_classifications.tex
\begin{CJK*}{UTF8}{gbsn}
\begin{table}[h]
\centering
\small
\begin{tabular}{lll}
\toprule
\textbf{Medical} & \textbf{Math} & \textbf{Remote Sensing} \\
\midrule
\texttt{ly}          & \texttt{1}          & \texttt{<id 220>}  \\
他 & \texttt{<id 220>}   & \texttt{...}       \\
\texttt{<id 220>}    & \texttt{2}          & \texttt{<id 4102>} \\
方面的       & \texttt{0}          & \texttt{/or}       \\
\texttt{/or}         & \texttt{(}          & \texttt{wards}     \\
\texttt{ness}        & \texttt{=}          & \texttt{\textbackslash n} \\
\bottomrule
\end{tabular}
\caption{\textbf{Representative noisy ExpertLens token classifications.}
Because \expertlens infers expert specialization by classifying decoded vocabulary
tokens, ambiguous tokens, token fragments, punctuation, and control tokens can
occasionally be assigned to a semantic domain. We show representative examples
for each domain.}
\label{tab:expertlens_noisy_tokens}
\end{table}
\end{CJK*}

%% file: prompts/EL_system.tex
\newtcblisting{psmall}{
  enhanced jigsaw,
  breakable,
  listing only,
  blank,
  borderline={1pt}{-5pt},
  listing options={
    breakindent=0pt,
    breaklines=true,
    basicstyle=\ttfamily\tiny,
    inputencoding=utf8,
    extendedchars=true,
    literate=
      {-}{{--}}1
      {--}{{---}}1
      {“}{{``}}1
      {”}{{''}}1
      {‘}{{`}}1
      {’}{{'}}1
  }
}

\begin{psmall}
Classify each token list below (one MoE expert per item).
Tokens may be English, Chinese, or other languages.
Include partial/subword matches (e.g. physio, histo, sat, geo, raster, ndvi, geom, digits, math symbols).

Return JSON only with one entry per expert below (same expert_id values, none omitted):
{ "items": [ { "expert_id": <int>, "labels": [[<specialist>, ...], ...] } ] }
Each inner list is one token's specialist tags (subset of medicine, geospatial, math).
Use [] for generalist-only tokens. A token may appear in multiple inner lists' domains.

Experts:
  expert_id=<expert ID> tokens=[<10 decoded tokens>]
  expert_id=<expert ID> tokens=[<10 decoded tokens>]
  ...
\end{psmall}
\captionof{figure}{\textbf{ExpertLens system prompt.}}
\label{prompts:system_prompt}

%% file: prompts/EL_classification.tex
\begin{psmall}
You classify vocabulary tokens from a MoE router logit lens (top predicted tokens per expert).

These tokens are often BPE or byte-level fragments, NOT only complete dictionary words.
Judge semantic association generously using morphology, substrings, symbols, and context.

For EACH token, return a list of specialist domains that apply (zero or more):
  medicine, geospatial, math

A token may match MULTIPLE specialists (e.g. a coordinate token can be both geospatial and math, or a biomedical measurement can be both medicine and math).
Use an empty list [] when the token is generalist only (no specialist signal).

NUMBERS, UNITS, AND SYMBOLS:
  - Digit / numeric tokens (0, 1, 2, 12, 0.5, ...): tag math when they indicate quantities, formulas, coordinates, measurements, or other numeric reasoning.
  - Percent sign (\%) and percent-related fragments: tag math.
  - Latitude/longitude, coordinate, map-scale, band-index, raster-size, pixel-size, ground-sampling-distance/GSD, resolution, and geodesic notation may be tagged both geospatial and math.

Do NOT use "generalist" inside per-token lists; use [] instead.

medicine -- clinical / biomedical / healthcare (patient, disease, diagnosis, treatment, anatomy, physiology, pathology, radiology, cells, DNA, gene, symptom, drug, hospital, ...).
geospatial -- geospatial / remote-sensing / satellite, aerial, overhead, drone, Earth-observation, GIS, mapping, terrain/topography, land cover/use, agriculture/crops, urban features, roads/streets/highways/intersections/bridges/railways/runways/airports, buildings/roofs/parking lots, water/lakes/ponds/rivers/streams/canals/reservoirs/wetlands/coastlines/beaches/ocean/sea, floods, wildfire/burn scars, clouds/shadows, elevation DEM/DSM/DTM, LiDAR/point clouds, SAR/InSAR/radar/backscatter, panchromatic/RGB/false-color/NIR/SWIR/multispectral/hyperspectral bands, Sentinel/Landsat/MODIS/Planet/Maxar/WorldView/NAIP, NDVI/EVI, raster/vector/tile/chip/scene/polygon/bounding box, GeoTIFF/GeoJSON/shapefile, latitude/longitude/projection/CRS/UTM/WGS84/georeferencing, and related fragments.
math -- equations, operators, geometry/algebra fragments (geom, trig, calc, ^, ...).

Tokens may be English, Chinese, or other languages.
Respond with valid JSON only.
\end{psmall}
\captionof{figure}{\textbf{ExpertLens classification prompt.}}
\label{prompts:system_prompt}